\documentclass[sigconf]{acmart}

\usepackage{booktabs} 
\usepackage{color,soul}
\usepackage{amsmath}
\usepackage{bm}
\usepackage{mathtools}
\usepackage{helvet}  
\usepackage{courier}  
\usepackage{url}  
\usepackage{graphicx}  
\usepackage{algorithm, algorithmic}
\usepackage{multirow}
\usepackage{subfig}
\usepackage{balance}

\DeclareMathOperator*{\argmax}{arg\,max}

\newcommand{\sysname}{OAHU}

\def\BibTeX{{\rm B\kern-.05em{\sc i\kern-.025em b}\kern-.08emT\kern-.1667em\lower.7ex\hbox{E}\kern-.125emX}}
    
\copyrightyear{2019}
\acmYear{2019} 
\setcopyright{iw3c2w3}
\acmConference[WWW '19]{Proceedings of the 2019 World Wide Web Conference}{May 13--17, 2019}{San Francisco, CA, USA}
\acmBooktitle{Proceedings of the 2019 World Wide Web Conference (WWW '19), May 13--17, 2019, San Francisco, CA, USA}
\acmPrice{}
\acmDOI{10.1145/3308558.3313503}
\acmISBN{978-1-4503-6674-8/19/05}

\begin{document}

%
\title{Towards Self-Adaptive Metric Learning On the Fly}

%
\author{Yang Gao}
\affiliation{%
  \institution{The University of Texas at Dallas}
  \streetaddress{800 W. Campbell Road}
  \city{Richardson}
  \state{Texas}
  \postcode{75080}
}
\email{yxg122530@utdallas.edu}

\author{Yi-Fan Li}
\affiliation{%
  \institution{The University of Texas at Dallas}
  \streetaddress{ 800 W. Campbell Road}
  \city{Richardson}
  \state{Texas}
  \postcode{75080}
}
\email{yli@utdallas.edu}

\author{Swarup Chandra}
\affiliation{%
  \institution{The University of Texas at Dallas}
  \streetaddress{ 800 W. Campbell Road}
  \city{Richardson}
  \state{Texas}
  \postcode{75080}
}
\email{swarup.chandra@utdallas.edu}

\author{Latifur Khan}
\affiliation{%
  \institution{The University of Texas at Dallas}
  \streetaddress{ 800 W. Campbell Road}
  \city{Richardson}
  \state{Texas}
  \postcode{75080}
}
\email{lkhan@utdallas.edu}

\author{Bhavani Thuraisingham}
\affiliation{%
  \institution{The University of Texas at Dallas}
  \streetaddress{ 800 W. Campbell Road}
  \city{Richardson}
  \state{Texas}
  \postcode{75080}
}
\email{bhavani.thuraisingham@utdallas.edu}

%
%
\renewcommand{\shortauthors}{Gao, et al.}

%
\begin{abstract}
Good quality similarity metrics can significantly facilitate the performance of many large-scale, real-world applications.
Existing studies have proposed various solutions to learn a Mahalanobis or bilinear metric in an online fashion by either restricting distances between similar (dissimilar) pairs to be smaller (larger) than a given lower (upper) bound or requiring similar instances to be separated from dissimilar instances with a given margin. 
However, these linear metrics learned by leveraging fixed bounds or margins may not perform well in real-world applications, especially when data distributions are complex. We aim to address the open challenge of ``Online Adaptive Metric Learning'' (OAML) for learning adaptive metric functions on-the-fly.
Unlike traditional online metric learning methods, OAML is significantly more challenging since the learned metric could be non-linear and the model has to be self-adaptive as more instances are observed. In this paper, we present a new online metric learning framework that attempts to tackle the challenge by learning a ANN-based metric with adaptive model complexity from a stream of constraints. In particular, we propose a novel Adaptive-Bound Triplet Loss (ABTL) to effectively utilize the input constraints, and present a novel Adaptive Hedge Update (AHU) method for online updating the model parameters. We empirically validates the effectiveness and efficacy of our framework on various applications such as real-world image classification, facial verification, and image retrieval.
\end{abstract}

%
%


%

%
\maketitle

\section{Introduction}
\label{Sec:introduction}
Learning a good quality similarity metric for images is critical to many large-scale web applications such as image classification, face verification, and image retrieval. These applications are widely used for user identity verification in mobile apps and image recommendation in multi-media search engines. Therefore, having a scalable and high-performance technique for learning a good image similarity metric is becoming increasingly important.
Various online metric learning (OML) solutions applying neural networks~\cite{LiGWZHS18,ChechikSSB10,JainKDG08,JinWZ09,breen2002image} have been proposed to learn a similarity metric from a stream of constraints. These constraints are typically collected and constructed from user clicks,  indicating the similarity and dissimilarity relations among images. Two kinds of constraints, i.e., pairwise and triplet constraints, are widely-accepted in previous studies. A pairwise constraint consists of two similar or dissimilar images, while a triplet constraint is of the form $(A, B, C)$, where image $A$ is similar to image $B$, but is dissimilar to image $C$. Compared to offline metric learning approaches, OML algorithms do not require the entire training data to be made available prior to the learning task and hence are more suitable for web applications.

Despite the success of existing OML approaches in many applications, some intrinsic drawbacks have been recently observed. First, most of the existing OML algorithms~\cite{LiGWZHS18,ChechikSSB10,JainKDG08,JinWZ09} are designed to learn a pre-selected linear metric (e.g., Mahalanobis distance) by solving a convex optimization problem. These simple metrics are unable to capture the non-linear semantic similarities among instances in complex application scenarios such as image retrieval with fine-grained sub-classes. 
Second, the performance of existing OML solutions depend heavily on the quality of input constraints, i.e., the true similarity or dissimilarity among example images (training data) provided by the user. These algorithms learn the metric model only from the ``hard'' constraints within the input and are biased towards them, thereby weakening the model robustness. Here, a constraint is ``hard'' if the dissimilar images in it is computed as similar by current metric and vice-versa.
Moreover, filtering out ``hard'' constraints from a large volume of candidates is computationally expensive, and is impossible to perform in a real-time setting. In addition, a real-world input stream usually contains limited number of ``hard'' constraints because we have little control on the source, i.e., the users.
As a result, existing OML approaches may suffer from low constraint utilization in practical applications, leading to poor performance. 

These intrinsic drawbacks result in the open challenge of "Online Adaptive Metric Learning" (OAML), i.e., how to learn a metric function with adaptive complexity as more constraints are observed in the stream,  while maintaining high constraint utilization during the continuous learning process. This metric function should have a dynamic hypothesis space, ranging from simple linear functions to highly non-linear functions. 

In this paper, we propose a novel framework that is able to learn a ANN-based metric from a stream of constraints with full constraint utilization, and more importantly, is able to dynamically adapt its model complexity when necessary. 
To achieve this, we first amend the common ANN architecture so that the output of every hidden layer is given as input to an independent metric embedding layer (MEL). Each MEL represents an embedding space where similar instances are closer to each other and dissimilar instances are further away from each other.
A metric weight is then assigned to each MEL, measuring its importance in the entire metric model. 
We propose a novel Adaptive-Bound Triplet Loss (ABTL) to evaluate the similarity and dissimilarity errors introduced by the arriving constraint at each online round, and present a novel Adaptive Hedge Update (AHU) method to update the parameters of the metric model along with the metric weights based on these errors.
Note that the proposed ABTL eliminates the unique dependency on ``hard'' constraints in existing methods, thereby improving both the constraint utilization and the quality of learned metrics. 
We refer to the proposed approach as \underline{O}nline metric learning with \underline{A}daptive \underline{H}edge \underline{U}pdate (OAHU).

The contribution of this paper are as follows:
\begin{itemize}
    \item We present a novel online metric learning framework - called \sysname{} - that addresses the challenges in ``Online Adaptive Metric Learning'' by learning a ANN-based metric with adaptive model complexity and achieving full constraint utilization.
    \item We theoretically show the regret bound of \sysname{} and demonstrate the optimal range for parameter selection.
    \item We empirically evaluate \sysname{} on real-world image classification, face verification, and image retrieval tasks, and compare its results with existing state-of-the-art online metric learning algorithms.
    
\end{itemize}

Rest of this paper is organized as follows. We first review related works in Section~\ref{Sec:background}. Then, we formalize the online metric learning problem and discuss our proposed solution in Section~\ref{Sec:approach}. Next, we discuss the empirical evaluation of \sysname{} in Section~\ref{Sec:evaluation}, and finally conclude in Section~\ref{Sec:conclusion}.

\section{Related Work}
\label{Sec:background}
Existing online metric learning solutions can be broadly classified into two categories: bilinear similarity-based or Mahalanobis distance-based methods. 

The bilinear similarity-based methods learn a distance metric given by $d(\bm{x_1}, \bm{x_2})=\bm{x_1}^TA\bm{x_2}$. A popular bilinear-based method is OASIS~\cite{ChechikSSB10}. It is an online dual approach with a passive-aggressive algorithm which computes a similarity metric to solve the image retrieval task. Sparse Online Metric Learning (SOML)~\cite{soml} is another technique which learns a metric by simplifying the full matrix $A$ by a diagnoal matrix $w$, i.e., $A=diag(w)$, to handle very high-dimensional cases. To reduce the memory and computational
cost, the Sparse Online Relative Similarity (SORS)~\cite{sors} method adopts an off-diagonal $l_1$ norm of $A$ as the sparse regularization to pursue a sparse model during the learning process. These three methods are based on triplet constraints where relative similarity/dissimilarity among images are specified. 

On the other hand, Mahalanobis distance-based methods focus on learning a distance metric given by $d(\bm{x_1}, \bm{x_2})=(\bm{x_1}-\bm{x_2})^TA(\bm{x_1}-\bm{x_2})$. Several algorithms have been proposed. 
Pseudo-metric Online Learning Algorithm (POLA)~\cite{Shalev-ShwartzSN04} introduces successive projection operations to learn the optimal metric.
Whereas, LEGO~\cite{JainKDG08} learns a Manalanobis metric based on LogDet regularization and gradient descent. RDML~\cite{JinWZ09} uses a regularized online metric learning algorithm. Recently, Li et al.~\cite{LiGWZHS18} introduced a closed-form solution called OPML, which adopts a one-pass triplet construction strategy to learn the optimal metric with a low computational cost. All these methods, except OPML, are based on pairwise constraints where the similarity or dissimilarity among pairs of images are specified. 

In general, triplet constraints are more effective than pairwise constraints~\cite{ChechikSSB10,WeinbergerBS05,ShawHJ11} because they allow simultaneous learning of similar and dissimilar relations.
Therefore, similar to OASIS and OPML, we focus on triplets to learn a high-quality metric. 
However, in contrast to all the previous online metric learning algorithms, our approach does not depend on the Mahalanobis or bilinear distance since their linearity property~\cite{BelletHS13} may limit expressiveness when exposed to complex data distributions in practical applications. Instead, we introduce a self-adaptive ANN-based metric that flexibly selects an appropriate metric function according to the scenario on which it is deployed. 

\begin{table*}
\centering
\resizebox{0.9\textwidth}{!}{%
\begin{tabular}{|ll|}
\hline
$\bm{x_t^{*}}$: Any anchor, positive or negative instance in a triplet. &  $E_l$: $l^{th}$ metric model in \sysname{}.\\
$f^{(l)}(\bm{x_t^{*}})$: Metric embedding of $\bm{x_t^*}$ generated by $E_l$. & $\mathcal{L}^{(l)}(\bm{x_t}, \bm{x_t^{+}}, \bm{x_t^{-}})$: The local loss of $(\bm{x_t}, \bm{x_t^{+}}, \bm{x_t^{-}})$ evaluated by $E_l$. \\
$\mathcal{L}_{overall}(\bm{x_t}, \bm{x_t^{+}}, \bm{x_t^{-}})$: The total loss of $(\bm{x_t}, \bm{x_t^{+}}, \bm{x_t^{-}})$ evaluated in \sysname{}. & $\alpha^{(l)}$: Weight of metric model $E_l$. \\
$D^{(l)}(\bm{x_i}, \bm{x_j})$: Distance between $\bm{x_i}$ and $\bm{x_j}$ measured in $E_l$ based on $f^{(l)}(\bm{x})$. &  $d_{sim}^{(l)}(\bm{x_i}, \bm{x_j})$ / $d_{dis}^{(l)}(\bm{x_i}, \bm{x_j})$: Similarity/Dissimilarity threshold in $E_l$ for a pair $(\bm{x_i}, \bm{x_j})$.\\
$\mathcal{T}_{sim}^{(l)}$ / $\mathcal{T}_{dis}^{(l)}$:  Upper bound of $d_{sim}^{(l)}(\bm{x_i}, \bm{x_j})$ / Lower bound of $d_{dis}^{(l)}(\bm{x_i}, \bm{x_j})$. & $\mathcal{L}^{(l)}_{attr}(\bm{x_t}, \bm{x_t^{+}})$ / $\mathcal{L}^{(l)}_{rep}(\bm{x_t}, \bm{x_t^{-}})$: Attractive/Repulsive loss of $(\bm{x_t}, \bm{x_t^{+}})$ / $(\bm{x_t}, \bm{x_t^{-}})$ evaluated in $E_l$. \\
$L$: Number of hidden layers in \sysname{}. & $d$: Dimensionality of input data. \\ 
$S_{hidden}$: Number of units in each hidden layer in \sysname{}. & $S_{emb}$: Dimensionality of learned metric embedding. \\
$\beta$: Discount factor for the update of $\alpha^{(l)}$. & $s$: Smooth factor for the update of $\alpha^{(l)}$.\\
$\eta$: Learning rate. & $W^{(l)}$ / $\Theta^{(l)}$: Parameter of $l^{th}$ hidden layer/Parameter of $l^{th}$ metric embedding layer. \\
\hline
\end{tabular}%
}
\caption{Commonly used symbols and terms}
\vspace{-4mm}
\label{tab:symbols}
\end{table*}

\section{OAHU}
\label{Sec:approach}

\subsection{Problem Setting}
\label{subsec:ps}
Given a sequence of triplet constraints $S=\lbrace(\bm{x_t}, \bm{x_t^{+}}, \bm{x_t^{-}})\rbrace_{t=1}^{T}$ that arrive sequentially, where $\lbrace\bm{x_t},\bm{x_t^{+}},\bm{x_t^{-}}\rbrace \in \mathcal{R}^d$, and $\bm{x_t}$ (anchor) is similar to $\bm{x_t^{+}}$ (positive) but is dissimilar to $\bm{x_t^{-}}$ (negative), the goal of online adaptive metric learning is to learn a model $\bm{F}: \mathcal{R}^d \mapsto \mathcal{R}^{d'}$ such that $||\bm{F}(\bm{x_t})-\bm{F}(\bm{x_t^{+}})||_2\ll||\bm{F}(\bm{x_t})-\bm{F}(\bm{x_t^{-}})||_2$. 

The overall aim of the task is to learn a metric model with adaptive complexity while maintaining a high constraint utilization rate. Here, the complexity of $\bm{F}$ needs to be adaptive so that its hypothesis space is automatically enlarged or shrinked as necessary.

In practical applications, the input triplet stream can be generated as follows. First, some seed triplets can be constructed from user clicks. Then, more triplets are formed by taking transitive closures\footnote{If $(\bm{x_1}, \bm{x_2})$ and $(\bm{x_1}, \bm{x_3})$ are two similar pairs, then $(\bm{x_2}, \bm{x_3})$ is a similar pair. If $(\bm{x_1}, \bm{x_2})$ and $(\bm{x_2}, \bm{x_3})$ are two similar pairs, then $(\bm{x_1}, \bm{x_3})$ is a similar pair. If $(\bm{x_1}, \bm{x_2})$ is a similar pair and $(\bm{x_1}, \bm{x_3})$ is a dissimilar pair,  then $(\bm{x_2}, \bm{x_3})$ is a dissimilar pair. If $(\bm{x_1}, \bm{x_2})$ is a similar pair and $(\bm{x_2}, \bm{x_3})$ is a dissimilar pair, then $(\bm{x_1}, \bm{x_3})$ is a dissimilar pair.} over these seed triplets. The generated triplets are inserted into the stream in chronological order of creation.
In this paper, we assume the existence of such triplet stream and focus on addressing the challenges of online metric learning in such scenarios.

\subsection{Overview}

\begin{figure*}[t]
\centering
\includegraphics[width=0.8\textwidth]{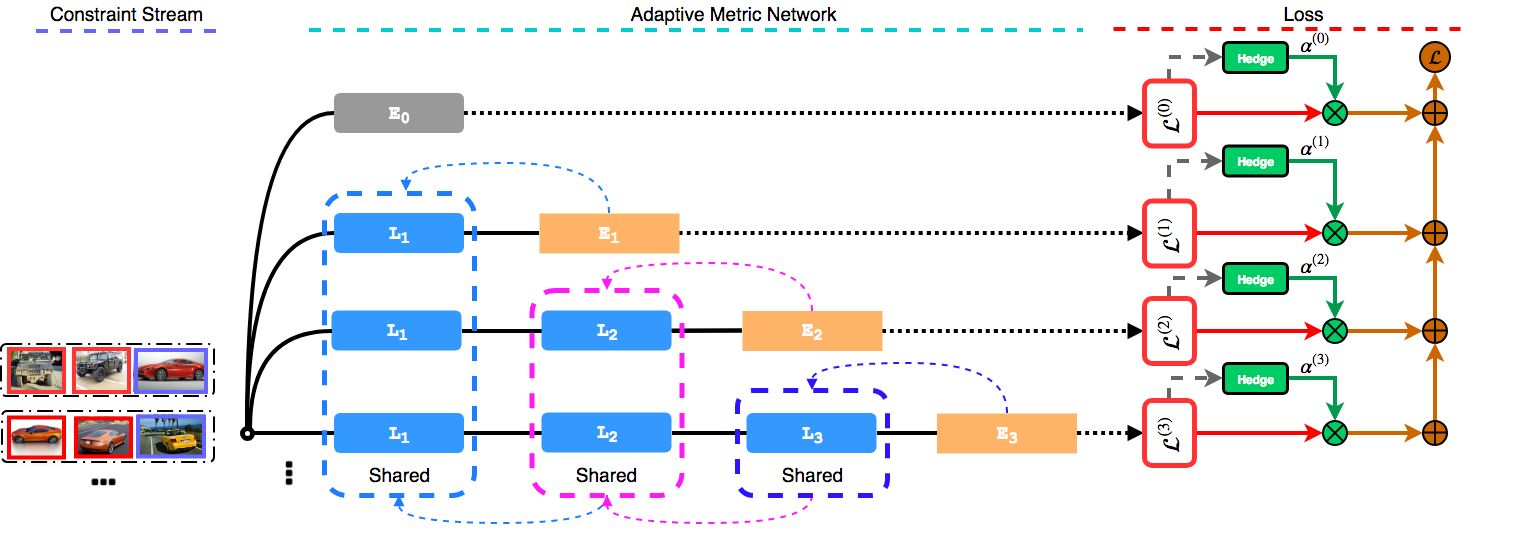}
\caption{Overview of the \sysname{}. The chromatic dashed arrows represent update contributions and directions. Each $L_i \in \lbrace L_1, L_2, \ldots \rbrace$ represents a linear transformation layer followed by a ReLU activation. $\lbrace E_0, E_1, \dots \rbrace$ are the embedding layers connected to corresponding input or hidden layers. Note that $E_0$ here represents a linear metric model, i.e., a linear transformation from the input feature space to the embedding space.}
\vspace{-2mm}
\label{fig:overview}
\end{figure*}

To learn a metric model with adaptive complexity, we need to address the following question: \textit{when} and \textit{how} to change the ``complexity'' of metric model in an \textit{online} setting? In this section, we will discuss the details of the proposed framework \sysname{} that addresses the question in a data-driven manner.

Figure~\ref{fig:overview} illustrates the structure of our \sysname{} framework. Inspired by recent works~\cite{SrinivasB16,HuangSLSW16}, we use an over-complete network and automatically adapt its effective depth to learn a metric function with an appropriate complexity based on input constraints. Consider a neural network with $L$ hidden layers:
we connect an independent embedding layer to the network input layer and each of these hidden layers. Every embedding layer represents a space where similar instances are closer to each other and vice-versa. Therefore, unlike conventional online metric learning solutions that usually learns a linear model, \sysname{} is an ensemble of models with varying complexities that share low-level knowledge with each other. For convenience, let $E_l \in \lbrace E_0, E_1, E_2, \ldots, E_L\rbrace$ denote the $l^{th}$ metric model in \sysname{}, i.e., the network branch starting from input layer to the $l^{th}$ metric embedding layer, as illustrated in Figure~\ref{fig:overview}. The simplest model in \sysname{} is $E_0$, which represents a linear transformation from the input feature space to the metric embedding space. A weight $\alpha^{(l)}\in [0,1]$ is assigned to $E_l$, measuring its importance in \sysname{}.

For a triplet constraint $(\bm{x_t}, \bm{x_t^{+}}, \bm{x_t^{-}})$ that arrives at time $t$, its metric embedding $f^{(l)}(\bm{x_t^*})$ generated by $E_l$ is

\small
\begin{equation}
\label{eq:metic_embedding}
    f^{(l)}(\bm{x_t^*}) = h^{(l)}\Theta^{(l)}
\end{equation}
\normalsize
where $h^{(l)} = \sigma(W^{(l)}h^{(l-1)}),$ with $l\ge 1, l\in \mathbb{N}$, and $h^{(0)} = \bm{x_t^*}$.

Here $\bm{x_t^*}$ denotes any anchor ($\bm{x_t}$), positive ($\bm{x_t^{+}}$) or negative ($\bm{x_t^{-}}$) instance based on the definition in Section~\ref{subsec:ps}, and $h^{(l)}$ represents the activation of $l^{\text{th}}$ hidden layer. 
Note that we also explicitly limit the learned metric embedding $f^{(l)}(\bm{x_t^*})$ to reside on a unit sphere, i.e., $||f^{(l)}(\bm{x_t^*})||_2=1$, to reduce the potential model search space and accelerate the  training of \sysname{}. 

During the training phase, for every arriving triplet $(\bm{x_t}, \bm{x_t^{+}}, \bm{x_t^{-}})$, \sysname{} first retrieves the metric embedding $f^{(l)}(\bm{x_t^*})$ from $l^{th}$ metric model according to Eq.~\ref{eq:metic_embedding}, and then produces a local loss $\mathcal{L}^{(l)}$ for $E_l$ by evaluating the similarity and dissimilarity errors based on $f^{(l)}(\bm{x_t^*})$. Thus, the overall loss introduced by this triplet is given by
\small
\begin{equation}
    \mathcal{L}_{overall}(\bm{x_t}, \bm{x_t^{+}}, \bm{x_t^{-}})=\sum\limits_{l=0}^L \alpha^{(l)} \cdot \mathcal{L}^{(l)}(\bm{x_t}, \bm{x_t^{+}}, \bm{x_t^{-}})
\end{equation}
\normalsize

Since the new architecture of \sysname{} introduces two sets of new parameters, $\Theta^{(l)}$ (parameters of $f^{(l)}$) and $\alpha^{(l)}$, we need to learn $\Theta^{(l)}$, $\alpha^{(l)}$ and $W^{(l)}$ during the online learning phase. Therefore, the final optimization problem to solve in \sysname{} at time $t$ is given by
\small
\begin{equation}
\begin{aligned}
& \underset{\Theta^{(l)},W^{(l)}, \alpha^{(l)}}{\text{minimize}}
& & \mathcal{L}_{overall} \\
& \text{subject to}
& & ||f^{(l)}(\bm{x_t^*})||_2 = 1, \forall l = 0, \ldots, L.
\end{aligned}
\end{equation}
\normalsize

To solve the above optimization problem, we propose a novel Adaptive-Bound Triplet Loss (ABTL) in Section~\ref{subsec:abel} to estimate $\mathcal{L}^{(l)}$ 
and introduce a novel Adaptive Hedge Update (AHU) method in Section~\ref{subsec:ahu} for online updating $\Theta^{(l)}$, $W^{(l)}$ and $\alpha^{(l)}$.

\subsection{Adaptive-Bound Triplet Loss}
\label{subsec:abel}

\subsubsection{Limitations of Existing Pairwise/Triplet Loss} 
\begin{figure}[t]
\centering
\includegraphics[width=0.5\columnwidth]{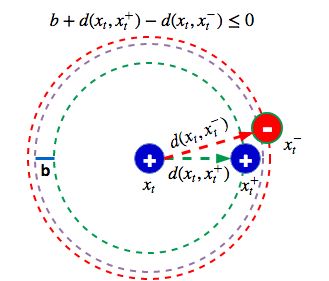}
\caption{A failure case of triplet loss.}
\vspace{-4mm}
\label{fig:failuremode}
\end{figure}

Existing online metric learning solutions typically optimize a pairwise or triplet loss in order to learn a similarity metric. A widely-accepted pairwise loss~\cite{Shalev-ShwartzSN04} and a common triplet loss~\cite{LiGWZHS18} are shown in Equation~\ref{eq:pairloss} and ~\ref{eq:tripletloss} respectively.  
\small
\begin{equation}
\label{eq:pairloss}
    L(x_t, x_t') = \max\lbrace0, y_{t}(d^2(x_{t}, x_{t}')-b)+1\rbrace
\end{equation}
\begin{equation}
\label{eq:tripletloss}
    L(x_t, x_t^+, x_t^-) = \max\lbrace0, b+d(x_t, x_t^+)-d(x_t, x_t^-)\rbrace
\end{equation}
\normalsize
Here $y_t\in\lbrace+1, -1\rbrace$ denotes whether $x_t$ is similar ($+1$) or dissimilar ($-1$) to $x_t'$ and $b \in \mathcal{R}$ is a user-specified fixed margin. Compared to the triplet loss that simultaneously evaluates both similarity and dissimilarity relations, the pairwise loss can only focus on one of the relations at a time, which leads to a poor metric quality. On the other hand, without a proper margin, 
the loose restriction of triplet loss can result in many failure cases. 
Here, a failure case is a triplet that produces a zero loss value, where dissimilar instances are closer than similar instances.
As the example shown in Figure~\ref{fig:failuremode}, for an input triplet $(x_t, x_t^+, x_t^-)$, although $x_t^+$ is 
closer to $x_t^-$ and further from $x_t$, models optimizing triplet loss would incorrectly ignore this constraint due to its zero loss value. In addition, selecting an appropriate margin is highly data-dependent, and requires extensive domain knowledge. Furthermore, as the model improves performance over time, a larger margin is required to avoid failure cases. 

Next, we describe an adaptive-bound triplet loss that attempts to solve these drawbacks.

\begin{figure}[t]
\centering
\includegraphics[width=0.7\columnwidth]{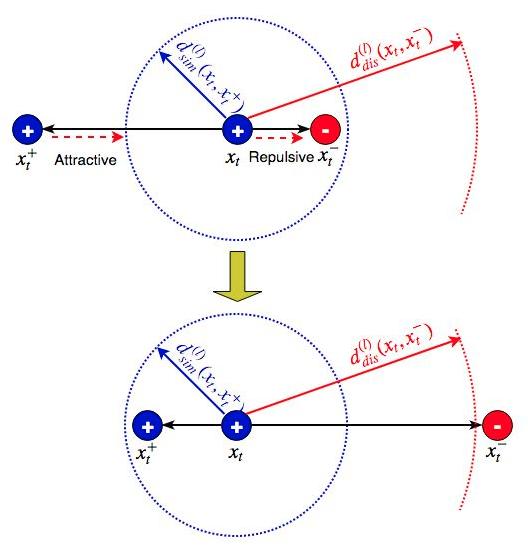}
\caption{Schematic illustration of ABTL.}
\vspace{-4mm}
\label{fig:abel}
\end{figure}

\subsubsection{Adaptive-Bound Triplet Loss (ABTL)} 
We define that instances of the same classes are mutually attractive to each other, while instances of different classes are mutually repulsive to each other. Therefore, for any input triplet constraint $(\bm{x_t}, \bm{x_t^{+}}, \bm{x_t^{-}})$, there is an attractive loss $\mathcal{L}_{attr}\in[0,1]$ between $\bm{x_t}$ and $\bm{x_t^{+}}$, and a repulsive loss $\mathcal{L}_{rep}\in[0,1]$ between $\bm{x_t}$ and $\bm{x_t^{-}}$. In the $l^{th}$ metric model $E_l \in \lbrace E_0, \ldots, E_L \rbrace$, we focus on a distance measure $D^{(l)}(\bm{x_i}, \bm{x_j})$ defined as follows:
\small
\begin{equation}
\label{eq:dist}
\begin{split}
D^{(l)}(\bm{x_i}, \bm{x_j}) =& ||f^{(l)}(\bm{x_i})-f^{(l)}(\bm{x_j})||_2, \quad \forall l=0,1,2,\ldots,L\\
\end{split}
\end{equation}
\normalsize
where $f^{(l)}(\bm{x})$ denotes the embedding generated by $E_l$. Since $||f^{(l)}(\bm{x})||_2$ is restricted to be $1$, we have $D^{(l)}(\bm{x_i}, \bm{x_j})\in [0,2]$.
In each iteration, let $D^{(l)}_{orig}(\bm{x_1},\bm{x_2})$ denote the distance between $\bm{x_1}$ and $\bm{x_2}$ before applying the update to $f^{(l)}$ (i.e., updating $\Theta^{(l)}$ and $\lbrace W^{(j)}\rbrace_{j=0}^l$) and $D^{(l)}_{update}(\bm{x_1},\bm{x_2})$ denote the distance after applying the update. 

Figure~\ref{fig:abel} illustrates the main idea of ABTL. The objective is to have the distance $D^{(l)}_{update}(\bm{x_t}, \bm{x_t^{+}})$ of two similar instances $\bm{x_t}$ and $\bm{x_t^{+}}$ to be less than or equal to a similarity threshold $d^{(l)}_{sim}(\bm{x_t}, \bm{x_t^{+}})$ so that the attractive loss $\mathcal{L}^{(l)}_{attr}(\bm{x_t}, \bm{x_t^{+}})$ drops to zero; on the other hand, for two dissimilar instances $\bm{x_t}$ and $\bm{x_t^{-}}$, we desire their distance $D^{(l)}_{update}(\bm{x_t}, \bm{x_t^{-}})$ to be greater than or equal to a dissimilarity threshold $d^{(l)}_{dis}(\bm{x_t}, \bm{x_t^{-}})$, thereby reducing the repulsive loss $\mathcal{L}^{(l)}_{rep}(\bm{x_t}, \bm{x_t^{-}})$ to zero. Equation~\ref{eq:constraint} presents these constraints mathematically.
\small
\begin{equation}
\label{eq:constraint}
    \left\{\begin{matrix}
    D^{(l)}_{update}(\bm{x_t}, \bm{x_t^{+}}) \le d^{(l)}_{sim}(\bm{x_t}, \bm{x_t^{+}})\\ 
    \\
    D^{(l)}_{update}(\bm{x_t}, \bm{x_t^{-}}) \ge d^{(l)}_{dis}(\bm{x_t}, \bm{x_t^-})
    \end{matrix}\right.
\end{equation}
\normalsize

Unfortunately, directly finding an appropriate $d^{(l)}_{sim}(\bm{x_t}, \bm{x_t^{+}})$ or $d^{(l)}_{dis}(\bm{x_t}, \bm{x_t^-})$ is difficult, since it varies with different input constraints. Therefore, we introduce a user-specified hyper-parameter $\tau>0$ to explicitly restrict the range of both thresholds.
Let $\mathcal{T}^{(l)}_{sim}$ denote the upper bound of $d^{(l)}_{sim}(\bm{x_t}, \bm{x_t^{+}})$, and $\mathcal{T}^{(l)}_{dis}$ denote the lower bound of $d^{(l)}_{dis}(\bm{x_t}, \bm{x_t^-})$, 

\small
\begin{equation}
    \left\{\begin{matrix}
    \mathcal{T}^{(l)}_{sim} = \tau \ge d^{(l)}_{sim}(\bm{x_t}, \bm{x_t^{+}})\\
    \\
    \mathcal{T}^{(l)}_{dis} = 2-\tau \leq d^{(l)}_{dis}(\bm{x_t}, \bm{x_t^-})
    \end{matrix}\right.
\end{equation}
\normalsize

Without loss of generality, we propose

\small
\begin{equation}
    \left\{\begin{matrix}
    d^{(l)}_{sim}(\bm{x_t}, \bm{x_t^{+}}) = a_1{\rm e}^{D^{(l)}_{orig}(\bm{x_t}, \bm{x_t^{+}})}+b_1\\ 
    \\
    d^{(l)}_{dis}(\bm{x_t}, \bm{x_t^-}) = -a_2{\rm e}^{-D^{(l)}_{orig}(\bm{x_t}, \bm{x_t^{-}})}+b_2
    \end{matrix}\right.
\end{equation}
\normalsize
where $a_1$, $b_1$, $a_2$ and $b_2$ are variables need to be determined.
Note that $d^{(l)}_{sim}(\bm{x_t}, \bm{x_t^{+}})$ and $d^{(l)}_{dis}(\bm{x_t}, \bm{x_t^-})$ are monotonically increasing and decreasing, respectively. Therefore, $d^{(l)}_{sim}$ preserves the information of original relative similarity, i.e., $d^{(l)}_{sim}(\bm{x_1}, \bm{x_2}) < d^{(l)}_{sim}(\bm{x_1}, \bm{x_3})$ if $D^{(l)}_{orig}(\bm{x_1}, \bm{x_2}) < D^{(l)}_{orig}(\bm{x_1}, \bm{x_3})$. This property is critical to many real-world applications that require a fine-grained similarity comparison, such as image retrieval.

The values of $a_1$, $b_1$, $a_2$ and $b_2$ are then determined by the following boundary conditions:
\small
\begin{equation}
\left\{\begin{matrix}
d^{(l)}_{sim}|_{D^{(l)}_{orig}(\bm{x_t},\bm{x_t^{+}})=2}\leq\mathcal{T}^{(l)}_{sim}=\tau &
d^{(l)}_{sim}|_{D^{(l)}_{orig}(\bm{x_t},\bm{x_t^{+}})=0}=0\\
d^{(l)}_{dis}|_{D^{(l)}_{orig}(\bm{x_t},\bm{x_t^-})=0}\ge \mathcal{T}^{(l)}_{dis}=2-\tau &
d^{(l)}_{dis}|_{D^{(l)}_{orig}(\bm{x_t},\bm{x_t^-})=2}=2
\end{matrix}\right.
\end{equation}
\normalsize

Thus we have
\small
\begin{equation}
    \left\{\begin{matrix}
    d^{(l)}_{sim}\big(\bm{x_t}, \bm{x_t^{+}}\big)=\frac{\tau}{{\rm e}^2-1}\big ({\rm e}^{D^{(l)}_{orig}(\bm{x_t}, \bm{x_t^{+}})}-1\big ) \\
    \\
    d^{(l)}_{dis}\big(\bm{x_t}, \bm{x_t^{-}}\big)= -\frac{2-(2-\tau)}{1-{\rm e}^{-2}}\big({\rm e}^{-D^{(l)}_{orig}(\bm{x_t}, \bm{x_t^{-}})}-1\big) +(2-\tau)
\end{matrix}\right.
\end{equation}
\normalsize

Provided with $d^{(l)}_{sim}$ and $d^{(l)}_{dis}$, the attractive loss $\mathcal{L}_{attr}$ and repulsive loss $\mathcal{L}_{rep}$ are determined by

\small
\begin{equation}
\label{eq:attr_rep}
    \left\{\begin{matrix}
    \mathcal{L}_{attr}^{(l)}(\bm{x_t}, \bm{x_t^{+}}) = \max \Big \lbrace 0,\frac{1}{2-c_1}D^{(l)}_{orig}(\bm{x_t}, \bm{x_t^{+}})-\frac{c_1}{2-c_1}\Big \rbrace\\ 
    \\
    \mathcal{L}_{rep}^{(l)}(\bm{x_t}, \bm{x_t^{-}}) = \max \Big \lbrace 0, \frac{-1}{c_2}D^{(l)}_{orig}(\bm{x_t}, \bm{x_t^-})+1 \Big \rbrace
    \end{matrix}\right.
\end{equation}
\normalsize
where $c_1=d^{(l)}_{sim}(\bm{x_t}, \bm{x_t^{+}})$ and $c_2=d^{(l)}_{dis}(\bm{x_t}, \bm{x_t^{-}})$.

Therefore, the local loss $\mathcal{L}^{(l)}(\bm{x_t}, \bm{x_t^{+}}, \bm{x_t^-})$, which measures the similarity and dissimilarity errors of the input triplet $(\bm{x_t}, \bm{x_t^{+}}, \bm{x_t^-})$ in $E_l$, is the average of $\mathcal{L}^{(l)}_{attr}$ and $\mathcal{L}^{(l)}_{rep}$:
\small
\begin{equation}
\label{eq:local_loss}
\begin{split}
    \mathcal{L}^{(l)}(\bm{x_t}, \bm{x_t^{+}}, \bm{x_t^-})&=\frac{1}{2}\Big(\mathcal{L}_{attr}^{(l)}(\bm{x_t}, \bm{x_t^{+}})+\mathcal{L}_{rep}^{(l)}(\bm{x_t}, \bm{x_t^{-}})\Big)\\
\end{split}
\end{equation}
\normalsize

Next, we answer an important question: What is the best value of $\tau$? Here, we demonstrate that the optimal range of $\tau$ is $(0, \frac{2}{3})$. The best value of $\tau$ is thus empirically selected within this range via cross-validation.

\begin{theorem}
\label{theo:1}
With $\tau \in (0, \frac{2}{3}) $, by optimizing the proposed adaptive-bound triplet loss, different classes are separated in the metric embedding space.
\end{theorem}
\begin{proof}
Let $D(c_1,c_2)$ denotes the minimal distance between classes $c_1$ and $c_2$, i.e., the distance between two closest instances from $c_1$ and $c_2$ respectively. Consider an arbitrary quadrupole $(x_1, x_2, x_3, x_4) \in \mathcal{Q}$ where $\lbrace x_1, x_2\rbrace \in c_1$, $\lbrace x_3,x_4 \rbrace \in c_2$, and $\mathcal{Q}$ is the set of all possible quadrupoles generated from class $c_1$ and $c_2$. Suppose  $(x_2,x_3)$ is the closest dissimilar pair among all possible dissimilar pairs that can be extracted from $(x_1, x_2, x_3, x_4)$ \footnote{it can always be achieved by re-arranging symbols of $x_1$,$x_2$,$x_3$ and $x_4$}. We first prove that the lower bound of $D(c_1,c_2)$ is given by $\min\limits_{(x_1, x_2, x_3, x_4)\in \mathcal{Q}} D^{(l)}(x_1, x_4)-D^{(l)}(x_1, x_2)-D^{(l)}(x_3, x_4)$.
Due to the triangle inequality property of p-norm distance, we have 
\small
\begin{equation}
\begin{split}
    D^{(l)}(x_1, x_4) &\leq D^{(l)}(x_1, x_2) + D^{(l)}(x_2, x_4)\\
    & \leq D^{(l)}(x_1, x_2) + D^{(l)}(x_2, x_3) + D^{(l)}(x_3, x_4)
\end{split}
\end{equation}
\normalsize
Therefore,
\small
\begin{equation}
\begin{split}
    D(c_1, c_2) &= \min_{(x_1, x_2, x_3, x_4)\in \mathcal{Q}} D^{(l)}(x_2, x_3) \\
    &\ge \min_{(x_1, x_2, x_3, x_4)\in \mathcal{Q}} D^{(l)}(x_1, x_4)-D^{(l)}(x_1, x_2)-D^{(l)}(x_3, x_4)
\end{split}
\end{equation}
\normalsize

By optimizing the adaptive-bound triplet loss, the following constraints are satisfied 

\small
\begin{equation}
    \left\{\begin{matrix}
    D^{(l)}(x_1, x_2) \le d^{(l)}_{sim}(x_1, x_2) \le \mathcal{T}^{(l)}_{sim}\\
    \\
    D^{(l)}(x_3, x_4) \le d^{(l)}_{sim}(x_3, x_4) \le \mathcal{T}^{(l)}_{sim}\\
    \\
    D^{(l)}(x_1, x_4) \ge d^{(l)}_{dis}(x_1, x_4) \ge \mathcal{T}^{(l)}_{dis}
    \end{matrix}\right.
\end{equation}
\normalsize

Thus
\small
\begin{equation}
\begin{split}
    D(c_1, c_2) 
    &\ge \min_{(x_1, x_2, x_3, x_4)\in \mathcal{Q}} D^{(l)}(x_1, x_4)-D^{(l)}(x_1, x_2)-D^{(l)}(x_3, x_4)\\
    &\ge \mathcal{T}^{(l)}_{dis} - 2\mathcal{T}^{(l)}_{sim}\\
    &=2-\tau-2\tau\\
    &=2-3\tau
\end{split}
\end{equation}
\normalsize
if $\tau \in (0, \frac{2}{3})$, we have $3\tau < 2$. Therefore,
\small
\begin{equation}
\label{eq:final_proof}
    D(c_1, c_2) \ge 2-3\tau > 0
\end{equation}
\normalsize
Equation~\ref{eq:final_proof} indicates that the minimal distance between class $c_1$ and $c_2$ is always positive so that these two classes are separated. 

\end{proof}

\subsubsection{Discussion}
Compared to pairwise and triplet loss, the proposed ABTL has several advantages:
\begin{itemize}
    \item Similar to triplet loss, ABTL simultaneously evaluates both similarity and dissimilarity errors to obtain a high-quality metric.
    \item The similarity threshold $d^{(l)}_{sim}(\bm{x_t}, \bm{x_t^{+}})$ is always less than or equal to $D^{(l)}_{orig}(\bm{x_t},\bm{x_t^{+}})$ and the dissimilarity threshold $d^{(l)}_{dis}(\bm{x_t}, \bm{x_t^{-}})$ is always greater than or equal to $D^{(l)}_{orig}(\bm{x_t},\bm{x_t^{-}})$. Thus, every input constraint contributes to model update, which eliminates the failure cases of triplet loss and leads to a higher constraint utilization rate.
    \item The optimal range of $\tau$ is provided with a theoretical proof.
\end{itemize}

\subsection{Adaptive Hedge Update (AHU)}
\label{subsec:ahu}
With adaptive-bound triplet loss, the overall loss in \sysname{} becomes
\small
\begin{equation}
\begin{split}
    \mathcal{L}_{overall}(\bm{x_t}, \bm{x_t^{+}}, \bm{x_t^{-}})&=\sum\limits_{l=0}^L \alpha^{(l)}\cdot\mathcal{L}^{(l)}(\bm{x_t}, \bm{x_t^{+}}, \bm{x_t^{-}})\\
    &=\sum\limits_{l=0}^L \frac{\alpha^{(l)}}{2}\Big(\mathcal{L}^{(l)}_{attr}(\bm{x_t}, \bm{x_t^{+}})+\mathcal{L}^{(l)}_{rep}(\bm{x_t}, \bm{x_t^{-}})\Big)
\end{split}
\end{equation}
\normalsize

We propose to learn $\alpha^{(l)}$ using the Hedge Algorithm~\cite{FreundS97}. Initially, all weights are uniformly distributed, i.e., $\alpha^{(l)}=\frac{1}{L+1}, l=0,1,2,\ldots, L$. At every iteration, for each metric model $E_l$, \sysname{} transforms the input triplet constraint into $E_l$'s metric embedding space, where the constraint similarity and dissimilarity errors are evaluated. Thus $\alpha^{(l)}$ is updated based on the local loss suffered by $E_l$ as:
\small
\begin{equation}
\label{eq:alpha_update}
    \alpha^{(l)}_{t+1} \leftarrow
    \begin{cases}
    \alpha^{(l)}_t\beta^{\mathcal{L}^{(l)}} & \beta^{\min\limits_{l}\mathcal{L}^{(l)}}\log\mathcal{L}^{(l)} > \beta-1\\
    \alpha^{(l)}_t[1-(1-\beta)\mathcal{L}^{(l)}]& otherwise
    \end{cases}
\end{equation}
\normalsize
where $\beta \in (0,1)$ is the discount factor. We choose to update $\alpha^{(l)}$ based on Equation~\ref{eq:alpha_update}, to maximize the update of $\alpha^{(l)}$ at each step. In other words, if a metric model produces a low/high $\mathcal{L}^{(l)}$ at time $t$, its associated weight $\alpha^{(l)}$ will gain as much increment/decrement as possible. Therefore, in \sysname{}, metric models with good performance are highlighted while models with poor performance are eliminated. 
Moreover, following Sahoo et.al~\cite{SahooPLH18}, to avoid the model bias issue~\cite{ChenGS15,GulcehreMVB16}, i.e., \sysname{} may unfairly prefer shallow models due to their faster convergence rate and excessively reduces weights of deeper models, we introduce a smooth factor $s$ used to set a minimal weight for each metric model. Thus, after updating weights according to Equation~\ref{eq:alpha_update}, the weights are set as: $\alpha^{(l)}_{t+1}=\max (\alpha^{(l)}_{t+1}, \frac{s}{L+1})$. 
Finally, at the end of every round, the weights $\alpha$ are normalized such that $\sum\limits_{l=0}^L \alpha^{(l)}_{t+1}=1$.

Learning the parameters $\Theta^{(l)}$ and $W^{(l)}$ is tricky but can be achieved via gradient descent. By applying Online Gradient Descent (OGD) algorithm, the update rules for $\Theta^{(l)}$ and $W^{(l)}$ are given by:
\small
\begin{equation}
\label{eq:theta_update}
\begin{split}
    \Theta^{(l)}_{t+1} &\leftarrow \Theta^{(l)}_{t} - \eta \nabla_{\Theta^{(l)}_t} \mathcal{L}_{overall}(\bm{x_t}, \bm{x_t^{+}}, \bm{x_t^{-}})\\
    &= \Theta^{(l)}_{t} - \eta \frac{\alpha^{(l)}}{2} \Big( \nabla_{\Theta^{(l)}_t}\mathcal{L}^{(l)}_{attr}(\bm{x_t}, \bm{x_t^{+}})+\nabla_{\Theta^{(l)}_t} \mathcal{L}^{(l)}_{rep}(\bm{x_t}, \bm{x_t^{-}})\Big)
\end{split}
\end{equation}
\begin{equation}
\label{eq:w_update}
    \begin{split}
    W^{(l)}_{t+1} &\leftarrow W^{(l)}_{t} - \eta \nabla_{W^{(l)}_t} \mathcal{L}_{overall}(\bm{x_t}, \bm{x_t^{+}}, \bm{x_t^{-}})\\
    &= W^{(l)}_{t} - \eta \sum\limits_{j=l}^L \frac{\alpha^{(j)}}{2} \Big( \nabla_{W^{(l)}_t}\mathcal{L}^{(j)}_{attr}(\bm{x_t}, \bm{x_t^{+}})+\nabla_{W^{(l)}_t} \mathcal{L}^{(j)}_{rep}(\bm{x_t}, \bm{x_t^{-}})\Big)
\end{split}
\end{equation}
\normalsize

Updating $\Theta^{(l)}$ (Equation~\ref{eq:theta_update}) is simple since $\Theta^{(l)}$ is specific to $E_l$, and is not shared across different metric models. In contrast, because deeper models share low-level knowledge with shallower models in \sysname{}, all models deeper than $l$ layers contribute to the update of $W^{(l)}$, as expressed in Equation~\ref{eq:w_update}. Algorithm~\ref{alg:\sysname{}} outlines the training process of \sysname{}.

\subsection{Regret Bound}
Overall, hedge algorithm enjoys a regret bound of $R_T\leq \sqrt{TlnN}$~\cite{FreundS97} where $N$ is the number of experts and $T$ is the number of trials. In our case, $N$ is the number of metric models (i.e., $L+1$) and $T$ is the number of input triplet constraints. The average worst-case convergence rate of \sysname{} is then given by $O(\sqrt{ln(N)/T})$. Therefore, \sysname{} is guaranteed to converge after learning from sufficiently many triplets.


\begin{algorithm}[t]
\caption{\textbf{\sysname{}: Online Metric Learning with Adaptive Hedge Update }}
\label{alg:\sysname{}}
\begin{algorithmic}[1]
\REQUIRE Discount Factor $\beta\in(0,1)$; Smooth Factor $s$; Control Parameter $\tau \in (0, \frac{2}{3})$; Learning rate $\eta$; A randomly initialized ANN with $L$ hidden layers that is parameterized by $\Theta^{(l)}$,$W^{(l)}$ and $\alpha^{(l)}$.
\ENSURE $\alpha^{(l)}$, $\Theta^{(l)}$ and $W^{(l)}$\\
\STATE Initialize $\alpha^{(l)}=\frac{1}{L+1}, \forall l=0,1,\ldots, L$ 
\FOR{$t=1,2,\ldots,T$}
\STATE Receive a triplet constraint $(\bm{x_t}, \bm{x_t^{+}}, \bm{x_t^-})$\\
\STATE Transform and retrieve the metric embedding $f^{(l)}(\bm{x_t})$, $f^{(l)}(\bm{x_t^{+}})$ and $f^{(l)}(\bm{x_t^-})$ from each model $E_l$.\\
\STATE Evaluate $\mathcal{L}^{(l)}_{attr}(\bm{x_t}, \bm{x_t^{+}})$ and $\mathcal{L}^{(l)}_{rep}(\bm{x_t}, \bm{x_t^{-}})$ (Eq.~\ref{eq:attr_rep}), $\forall l=0,1,\ldots,L$.
\STATE Compute $L^{(l)}(\bm{x_t}, \bm{x_t^{+}},\bm{x_t^{-}}), \forall l=0, 1, \ldots, L$ as per Eq.~\ref{eq:local_loss}.
\STATE Update $\Theta^{(l)}_{t+1},\forall l=0,1,2,\ldots,L$ as per Eq.~\ref{eq:theta_update}.\\
\STATE Update $W^{(l)}_{t+1},\forall l=0,1,2,\ldots,L$ as per Eq.~\ref{eq:w_update}.\\
\STATE Update $\alpha^{(l)}_{t+1},\forall l=0,1,2,\ldots,L$ as per Eq.~\ref{eq:alpha_update}.\\
\STATE $\alpha^{(l)}_{t+1}=\max (\alpha^{(l)}_{t+1}, \frac{s}{L+1}), \forall l=0,1,2,\ldots,L$.\\
\STATE Normalize $\alpha^{(l)}_{t+1}$, i.e., $\alpha^{(l)}_{t+1}=\frac{\alpha^{(l)}_{t+1}}{\sum\limits_{j=0}^L\alpha^{(j)}_{t+1}}, \forall l=0,1,2,\ldots,L$.
\ENDFOR

\end{algorithmic}
\end{algorithm}

\subsection{Application of \sysname{}}
\label{subsec:deployment}

In this part, we discuss the deployment of \sysname{} in different practical applications. Although \sysname{} can be applied in many applications, we mainly focus on the three most common tasks: classification, similarity comparison, and analogue retrieval. Note that the following discussion describes only one possible way of deployment for each task, and other deployment methods can exist.

\subsubsection{Classification}
\label{subsubsec:classification}
Given a training dataset $\mathcal{D}=\lbrace (\bm{x_i}, y_i)\rbrace_{i=1}^n$ where $\bm{x_i}\in \mathcal{R}^d$ is an training instance and $y_i\in \lbrace1,2,\ldots, c \rbrace$ is its associated label, we determine the label $y\in \lbrace1,2,\ldots,c\rbrace$ of a test instance $\bm{x}\in \mathcal{R}^d$ as follows:
\begin{itemize}
    \item For every metric model $E_l$ in \sysname{}, we find $k$ nearest neighbors of $\bm{x}$ measured by Eq.~\ref{eq:dist}, each of which is a candidate similar to $\bm{x}$. Hence, $k(L+1)$ candidates in total are found in \sysname{}. Let $\mathcal{N} =\lbrace \lbrace(\bm{x_j^{(l)}}, y^{(l)}_j)\rbrace_{j=1}^k \rbrace_{l=0}^L=\lbrace \mathcal{N}_l \rbrace_{l=0}^L$ denotes these candidates where $\bm{x_j^{(l)}}$ is the $j^{th}$ neighbor of $\bm{x}$ found in $E_l$ and $y^{(l)}_j$ is its associated label. The similarity score between $\bm{x}$ and $\bm{x_j^{(l)}}$ is given by 
    \small
    \begin{equation}
        S(\bm{x}, \bm{x_j^{(l)}})={\rm e}^{-\frac{D^{(l)}(\bm{x_j^{(l)}}, \bm{x})-d_{min}}{d_{max}-d_{min}}}\cdot \alpha^{(l)}
    \end{equation}
    \normalsize
    where
    $d_{min}=\min\limits_{ \mathcal{N}_l}D^{(l)}(\bm{x_j^{(l)}}, \bm{x})$ and
    $d_{max}=\max\limits_{ \mathcal{N}_l}D^{(l)}(\bm{x_j^{(l)}}, \bm{x})$.
    \item Then, the class association score of each class $c_i$ is calculated by:
    \small
    \begin{equation}
        S(c_i) = \sum\limits_{y_j^{(l)}=c_i} S(\bm{x}, \bm{x_j^{(l)}})
    \end{equation}
    \normalsize
    \item The final predicted label $y$ of $\bm{x}$ in \sysname{} is determined as:
    \small
    \begin{equation}
        y = \argmax\limits_{c_i} S(c_i)
    \end{equation}
    \normalsize
\end{itemize}
\subsubsection{Similarity Comparison}
The task of similarity comparison is to determine whether a given pair $(\bm{x_1}, \bm{x_2})$ is similar or not. To do this, a user-specified threshold $\mathcal{T}\in (0,1)$ is needed. The steps below are followed to perform the similarity comparison in \sysname{}.
\begin{itemize}
    \item First, for every metric model $E_l$, we compute the distance $D^{(l)}(\bm{x_1}, \bm{x_2})$ between $\bm{x_1}$ and $\bm{x_2}$. Since $D^{(l)}(\bm{x_1}, \bm{x_2}) \in [0,2]$, we divide $D^{(l)}(\bm{x_1}, \bm{x_2})$ by $2$ for normalization.
    \item The similarity probability of $(\bm{x_1}, \bm{x_2})$ is then determined by
    \small
    \begin{equation}
    \label{eq:sim_com}
        P(\bm{x_{1}},\bm{x_{2}}) = \sum\limits_{l=0}^L \alpha^{(l)}\cdot p_{l}
    \end{equation}
    \normalsize
    where $p_{l}=1$ if $D^{(l)}(\bm{x_1}, \bm{x_2})/2<\mathcal{T}$; Otherwise, $p_l=0$.
    \item $(\bm{x_1},\bm{x_2})$ is similar if $ P(\bm{x_1}, \bm{x_2})\ge 0.5$ and is dissimilar if $ P(\bm{x_1}, \bm{x_2})< 0.5$.
\end{itemize}
\subsubsection{Analogue Retrieval}
Given a query item $\bm{x}$, the task of analogue retrieval is to find $k$ items that are most similar to $\bm{x}$ in a database $\mathcal{D}$. Note that analogue retrieval is a simplified ``classification'' task where we only find similar items and ignore the labeling issue.
\begin{itemize}
\item We first retrieve the $k(L+1)$ candidates that are similar to $\bm{x}$ following the same process in classification task. Let $\mathcal{N} =\lbrace \lbrace \bm{x_j^{(l)}}\rbrace_{j=1}^k \rbrace_{l=0}^L=\lbrace \mathcal{N}_l \rbrace_{l=0}^L$ denote these candidates. The similarity score between $\bm{x_j^{(l)}}$ and $\bm{x}$ is:
\small
\begin{equation}
\label{eq:retrieval}
    Sim(\bm{x_j^{(l)}}, \bm{x}) = {\rm e}^{-\frac{D^{(l)}(\bm{x_j^{(l)}}, \bm{x})-d_{min}}{d_{max}-d_{min}}}\cdot \alpha^{(l)}
\end{equation}
\normalsize
where $d_{min}=\min\limits_{ \mathcal{N}_l}D^{(l)}(\bm{x_j^{(l)}}, \bm{x})$ and $d_{max}=\max\limits_{ \mathcal{N}_l}D^{(l)}(\bm{x_j^{(l)}}, \bm{x})$.
\item Then we sort these items in descending order based on their similarity scores and keep only the top $k$ distinct items as the final retrieval result.
\end{itemize}

\subsection{Time and Space Complexity}
Overall, the execution overhead of \sysname{} mainly arises from the training of the ANN-based metric model, especially from the gradient computation. Assume that the maximum time complexity for calculating the gradient of one layer is a constant $C$. The overall time complexity of \sysname{} is simply $O(nL^2C)$, where $n$ is the number of input constraints and $L$ is the number of hidden layers. Therefore, with pre-specified $L$ and ANN architecture, \sysname{} is an efficient online metric learning algorithm that runs in linear time.  Let $d$, $S_{hidden}$ and $S_{emb}$ denote the dimensionality of input data, number of units in each hidden layer and metric embedding size respectively. The space complexity of \sysname{} is given by $O\big(d\cdot S_{emb} + L\cdot S_{hidden}(d+S_{emb})+\frac{L(L-1)}{2}S_{hidden}^2\big)$.

\subsection{Extendability}
As discussed in Section~\ref{subsec:deployment}, \sysname{} is a fundamental module that can perform classification, similarity comparison, or analogue retrieval tasks independently. Therefore, it can be easily plugged into any existing deep models (e.g., a CNN or RNN) by replacing the classifier, i.e., fully-connected layers, with \sysname{}. The parameters of those feature-extraction layers (e.g., a convolutional layer) can be fine-tuned by optimizing $\mathcal{L}_{overall}$ directly. However, in this paper, to prove the validity of the proposed approach, we only focus on \sysname{} itself and leave the deep model extension for future work.

\section{Empirical Evaluation}
\label{Sec:evaluation} 
To verify the effectiveness of our method, we evaluate \sysname{} on three typical tasks, including (1) image classification (\textit{\textbf{classification}}), (2) face verification (\textit{\textbf{similarity comparison}}), and (3) image retrieval (\textit{\textbf{analogue retrieval}}).

\subsection{Baselines}
    To examine the quality of metrics learned in \sysname{}, we compare it with several state-of-the-art online metric learning algorithms. (1) LEGO~\cite{JainKDG08} (pairwise constraint): an online metric learning algorithm that learns a Mahalanobis metric based on LogDet regularization and gradient descent. (2) RDML~\cite{JinWZ09} (pairwise constraint): an online algorithm for regularized distance metric learning that learns a Mahalanobis metric with a regret bound. (3) OASIS~\cite{ChechikSSB10} (triplet constraint): an online algorithm for scalable image similarity learning that learns a bilinear similarity measure over sparse representations. Finally, (4) OPML~\cite{LiGWZHS18} (triplet constraint): a one-pass closed-form solution for online metric learning that learns a Mahalanobis metric with a low computational cost. 

\subsection{Implementation}
We have implemented \sysname{} using Python 3.6.2 and Pytorch 0.4.0 library. All baseline methods were based on code released by corresponding authors, except LEGO and RDML. Due to the unavailability of a fully functional code of LEGO and RDML, we use our own implementation based on the authors' description~\cite{JainKDG08,JinWZ09}. Hyper-parameter setting of these approaches, including \sysname{}, varies with different tasks, and will be discussed later for each specific task.

\subsection{Image Classification}
\subsubsection{Dataset and Classifier} 
To demonstrate the validacy of \sysname{} on complex practical applications, we use three publicly available real-world benchmark image datasets, including \textbf{FASHION-MNIST}\footnote{https://github.com/zalandoresearch/fashion-mnist}~\cite{online}, \textbf{EMNIST}\footnote{https://www.nist.gov/itl/iad/image-group/emnist-dataset}~\cite{CohenATS17} and \textbf{CIFAR-10}\footnote{https://www.cs.toronto.edu/~kriz/cifar.html}~\cite{Krizhevsky09learningmultiple} for evaluation. To be consistent with other image datasets, we convert images of CIFAR-10 into grayscale through the OpenCV API~\cite{opencv_library}, resulting in 1024 features. 
The \textit{k-NN} classifier is employed here, since it is widely used for classification tasks with only one parameter, and is also consistent with the deployment of \sysname{} discussed in Section~\ref{subsubsec:classification}. 
Details of these datasets are listed in ``Image Classification'' part of Table~\ref{tab:dataset}.

\begin{table}[t]
\centering
\resizebox{0.8\columnwidth}{!}{%
\begin{tabular}{|c|c|r|r|r|}
\hline
\textbf{Task} & \textbf{Dataset} & \multicolumn{1}{c|}{\textbf{\# features}} & \multicolumn{1}{c|}{\textbf{\# classes}} & \multicolumn{1}{c|}{\textbf{\# instances}} \\ \hline
\multirow{3}{*}{\textbf{Image Classification}} & FASHION-MNIST & 784 & 10 & 70,000 \\ \cline{2-5} 
 & EMNIST & 784 & 47 & 131,600 \\ \cline{2-5} 
 & CIFAR-10 & 1024 & 10 & 60,000 \\ \hline
\textbf{Face Verification} & LFW & 73 & 5749 & 13,233 \\ \hline
\multirow{2}{*}{\textbf{Image Retrieval}} & CARS-196 & 4096 & 196 & 16,185 \\ \cline{2-5} 
 & CIFAR-100 & 4096 & 100 & 60,000 \\ \hline
\end{tabular}%
}
\caption{Description of Datasets}
\vspace{-4mm}
\label{tab:dataset}
\end{table}

\begin{table*}[t]
\centering
\resizebox{0.8\textwidth}{!}{%
\begin{tabular}{|c|l|l|l|l|l|l|l|l|l|c|}
\hline
\multirow{2}{*}{\textbf{Methods}} & \multicolumn{3}{c|}{\textbf{FASHION-MNIST}} & \multicolumn{3}{c|}{\textbf{EMNIST}} & \multicolumn{3}{c|}{\textbf{CIFAR-10}} & \multicolumn{1}{c|}{\multirow{2}{*}{\textbf{win/tie/loss}}} \\ \cline{2-10}
 & \multicolumn{1}{c|}{\textit{Error Rate}} & \multicolumn{1}{c|}{$F_1$} & \multicolumn{1}{c|}{$\mathcal{U}$} & \multicolumn{1}{c|}{\textit{Error Rate}} & \multicolumn{1}{c|}{$F_1$} & \multicolumn{1}{c|}{$\mathcal{U}$}  & \multicolumn{1}{c|}{\textit{Error Rate}} & \multicolumn{1}{c|}{$F_1$} &  \multicolumn{1}{c|}{$\mathcal{U}$} & \multicolumn{1}{c|}{} \\ \hline
\textbf{LEGO} & 0.35$\pm$0.01 $\bullet$ & 0.64$\pm$0.00 & 0.58$\pm$0.00 & 0.84$\pm$0.00 $\bullet$ & 0.14$\pm$0.02 & 0.66$\pm$0.01 & 0.88$\pm$0.00 $\bullet$ & 0.05$\pm$0.02 & 0.59$\pm$0.00 & \textbf{3/0/0} \\ \hline
\textbf{RDML} & 0.22$\pm$0.01 $\bullet$ & 0.78$\pm$0.00 & 0.47$\pm$0.01 & 0.64$\pm$0.03 $\bullet$ & 0.32$\pm$0.05 & 0.50$\pm$0.00 & 0.76$\pm$0.01 $\bullet$ & 0.24$\pm$0.01 & 0.50$\pm$0.01 & \textbf{3/0/0} \\ \hline
\textbf{OASIS} & 0.21$\pm$0.02 $\bullet$ & 0.79$\pm$0.01 & 0.14$\pm$0.00 & 0.41$\pm$0.01 $\bullet$ & 0.59$\pm$0.00 & 0.24$\pm$0.00 & 0.79$\pm$0.02 $\bullet$ & 0.20$\pm$0.03 & 0.47$\pm$0.00 & \textbf{3/0/0} \\ \hline
\textbf{OPML} & 0.23$\pm$0.00 $\bullet$ & 0.78$\pm$0.00 & 0.30$\pm$0.01 & 0.43$\pm$0.01 $\bullet$ & 0.51$\pm$0.02 & 0.45$\pm$0.00 & 0.83$\pm$0.01 $\bullet$ & 0.16$\pm$0.00 & 0.48$\pm$0.04 & \textbf{3/0/0} \\ \hline
\textbf{\sysname{}} & \textbf{0.18$\pm$0.00} & \textbf{0.81$\pm$0.00} &
\textbf{1.00$\pm$0.00} &
\textbf{0.35$\pm$0.01} & \textbf{0.64$\pm$0.01} &
\textbf{1.00$\pm$0.00} & 
\textbf{0.68$\pm$0.00} & \textbf{0.31$\pm$0.00} &
\textbf{1.00$\pm$0.00} &
 - \\ \hline
\end{tabular}%
}
\caption{Comparison of classification performance on competing methods over benchmark image datasets. $\bullet/\circ$ indicates \textbf{\sysname{}} performs statistically better/worse (0.05 significance level) than the respective method according to the $p$-values. The statistics of win/tie/loss is also included. Both mean and standard deviation of error rates, constraint utilization $\mathcal{U}$ and macro $F_1$ scores are reported. $0.00$ denotes a value less than $0.005$.}
\vspace{-6mm}
\label{tab:classification}
\end{table*}

\begin{figure*}[t]
\centering
\subfloat[FASHION-MNIST\label{fig:fashionmnist_triplet}]{\includegraphics[width =0.3 \textwidth]{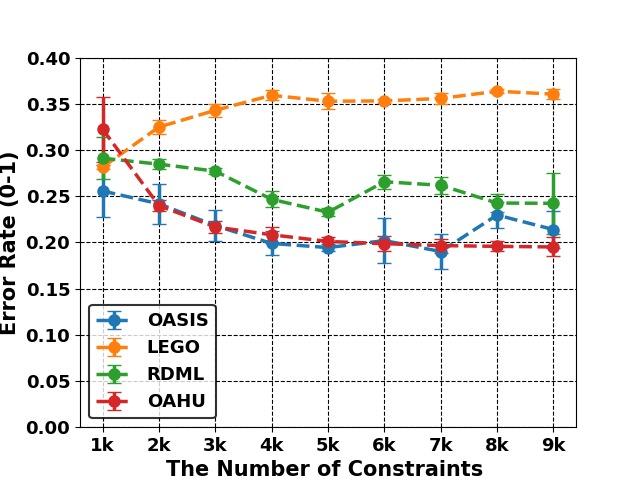}}\hfill
\subfloat[CIFAR-10\label{fig:cifar10_triplet}]{\includegraphics[width = 0.3\textwidth]{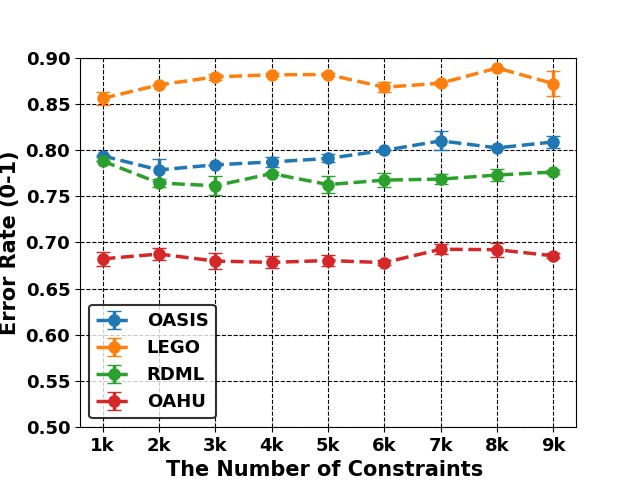}}
\hfill
\subfloat[\label{fig:weight_triplet}]{\includegraphics[width = 0.3\textwidth]{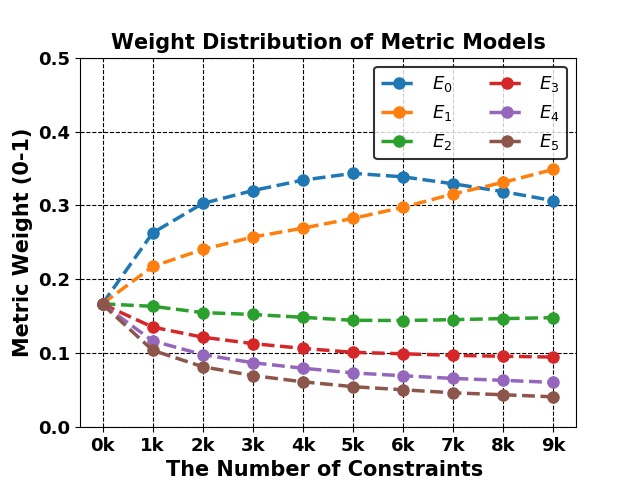}}
\caption{(a) / (b) Error rates of competing methods with increasing number of constraints on FASHION-MNIST / CIFAR-10 dataset. (c) The evolvement of metric weight distribution in \sysname{} with increasing number of constraints on FASHION-MNIST dataset. }
\vspace{-4mm}
\label{fig:error_rate_vary_constraints}
\end{figure*}

\subsubsection{Experiment Setup and Constraint Generation}
\label{subsubsec: escg}
To perform the experiment, we first randomly shuffle each dataset and then divide it into development and test sets with the split ratio of $1:1$. Hyper-parameters of these baseline approaches were set based on values reported by the authors and fine-tuned via 10-fold cross-validation on the development set. Because OPML applies a one-pass triplet construction process to generate the constraints, we simply provide the training data in the development set as the input for OPML.
For other approaches including \sysname{}, we first randomly sample $5,000$ seeds (i.e., pairwise or triplet constraints) from the training data and then construct $5,000$ more constraints by taking transitive closure over these seeds.
Note that the same number of constraints were adopted by the authors of LEGO~\cite{JainKDG08} and OPML~\cite{LiGWZHS18} when comparing their approaches with baselines. Thus our comparison is fair. 
In \sysname{}, we set $\beta=0.99$, $L=5$, $S_{hidden}=100$, $s=0.1$, $\eta=0.3$ and $S_{emb}=50$ as default. The value of $\tau$ was found by 10-fold cross-validation on the development set.
To reduce any bias that may result from random partition, all the classification results are averaged over $10$ individual runs with independent random shuffle processes.

\subsubsection{Evaluation Metric}
Similar to most of the baselines, e.g., LEGO~\cite{JainKDG08} and OPML~\cite{LiGWZHS18}, we adopt the error rate and macro $F_1$ score\footnote{http://scikit-learn.org/stable/modules/generated/sklearn.metrics.f1\_score.html} as the evaluation criterion. Moreover, the $p$-values of student's t-test were calculated to check statistical significance.  In addition, the statistics of win/tie/loss are reported according to the obtained p-values. The constraint utilization $\mathcal{U}$, i.e., the fraction of input constraints that actually contribute to model update, is also reported for each method.

\subsubsection{Analysis}
Table~\ref{tab:classification} compares the classification performance of \sysname{} with baseline methods on benchmark datasets. We observe that LEGO and RDML perform poorly on complex image datasets, e.g., EMNIST ($47$ classes), since they learn a Mahalanobis metric from pairwise constraints, and are biased to either the similarity or dissimilarity relation when updating model parameters. OASIS performs better than OPML because the bilinear metric eliminates the positive and symmetric requirement of the covariance matrix, i.e., $A$ in $d(\bm{x_1}, \bm{x_2})=\bm{x_1}^TA\bm{x_2}$ which results in a larger hypothesis space~\cite{ChechikSSB10}. However, \sysname{} outperforms all the baseline approaches by
providing not only significantly lower error rate, but also higher $F_1$ score, indicating the best overall classification performance among all competing methods.

To illustrate the effect of different numbers of constraints on the learning of a metric, we vary the  input constraint numbers\footnote{It is done by first generating $10,000$ constraints as discussed in Section~\ref{subsubsec: escg} and then uniformly sample the desired number of constraints from them.} for all competing methods except OPML. Here we do not compare with OPML because it internally determines the number of triplets to use via a one-pass triplet construction process. Therefore, explicitly varying the input triplet number for OPML is impossible. Figure~\ref{fig:fashionmnist_triplet} and Figure~\ref{fig:cifar10_triplet} show the comparison of error rates on FASHION-MNIST and CIFAR-10 dataset respectively. 
The large variance of error rates observed in both OASIS and RDML (i.e., error bars in the figures) indicates the strong correlation of their classification performance with the quality of input constraints.
All baseline approaches present an unstable, and sometimes, even worse performance with an increasing number of constraints. This is because these baselines are biased to those "hard" constraints leading to positive loss values and incorrectly ignore the emerging failure cases (see Figure~\ref{fig:failuremode}). In contrast, \sysname{} provides the best and most stable classification performance with minimal variance, indicating its robustness to the quality of input constraints and better capability of capturing the semantic similarity and dissimilarity relations in data. 

\sysname{} is better mainly because it not only learns from every input constraint by optimizing the adaptive-bound triplet loss (see constraint utilization $\mathcal{U}$ in Table~\ref{tab:classification}), but also automatically adapts the model complexity to achieve better performance (see Figure~\ref{fig:weight_triplet}).

\subsection{Face Verification}
\subsubsection{Dataset and Experiment Setup}
For face verification, we evaluate our method on Labeled Faces in the Wild Database (LFW)~\cite{LFWTechUpdate}. 
This dataset has two views: View 1 is used for development purpose (contains a training and a test set), and View 2 is taken as a benchmark for comparison (i.e., a 10-fold cross validation set containing pairwise images). 
The goal of face verification in LFW is to determine whether a pair of face images belongs to the same person. 
Since OASIS, OPML, and \sysname{} are triplet-based methods, following the setting in OPML~\cite{LiGWZHS18}, we adopt the image unrestricted configuration to conduct experiments for a fair comparison. 
In the image unrestricted configuration, the label information (i.e. actual names) in training data can be used and as many pairs or triplets as one desires can be formulated. We use View 1 for hyper-parameter tuning and evaluate the performance of all competing algorithms on each fold (i.e., 300 intra-person and 300 inter-person pairs) in View 2. Note that the default parameter values in \sysname{} and baselines were set using the same values as discussed in Section~\ref{subsubsec: escg}. Attribute features of LFW are used in the experiment so that the result can be easily reproduced and the comparison is fair. Attribute features\footnote{http://www.cs.columbia.edu/CAVE/databases/pubfig/download/lfw\_attributes.txt} ($73$ dimension) provided by Kumar et al.~\cite{KumarBBN09} are ``high-level'' features describing nameable attributes such as race, age etc., of a face image. Details of LFW are described in ``Face Verification'' part in Table~\ref{tab:dataset}.

\subsubsection{Constraint Generation and Evaluation Metric}
We follow the same process described in Section~\ref{subsubsec: escg} to generate constraints for training purpose. However, we add an additional step here: if a generated constraint contains any pair of instances in the validation set (i.e., all pairs in View 1) or the test set (i.e., all folds in View 2), we simply discard it and re-sample a new constraint. This step is repeated until the new constraint does not contain any pair in both sets. In the experiment, $10,000$ pair or triplet constraints in total are generated for training. 

Note, due to the one-pass triplet generation strategy of OPML, we let itself determine the training triplets.
For each test image pair, we first compute the distance between them measured by the learned metric, and then normalize the distance into $[0,1]$. The widely accepted Receiver Operating Characteristic (ROC) curve and Area under Curve (AUC) are adopted as the evaluation metric. 

\subsubsection{Analysis}

\begin{figure}[t]
\centering
\subfloat[\label{fig:roc}]{\includegraphics[width =0.5 \columnwidth]{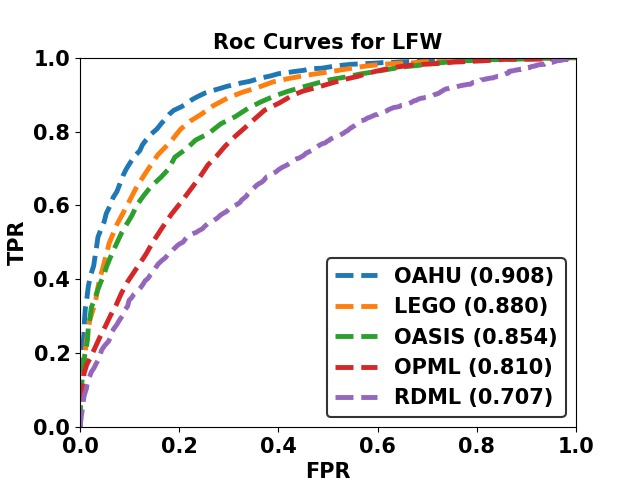}}\hfill
\subfloat[\label{fig:roc_weight}]{\includegraphics[width = 0.5\columnwidth]{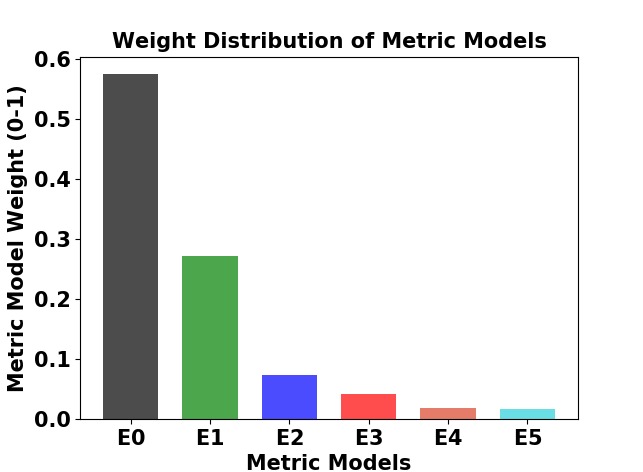}}
\caption{(a) ROC Curves of \sysname{} and contrastive methods on LFW dataset. AUC value of each method is given in bracket. (b) Metric weight distribution of \sysname{}.}
\vspace{-4mm}
\label{fig:roc_combine}
\end{figure}

\begin{figure}[t]
\centering
\subfloat[Failure Cases]{\includegraphics[width = 0.49\columnwidth]{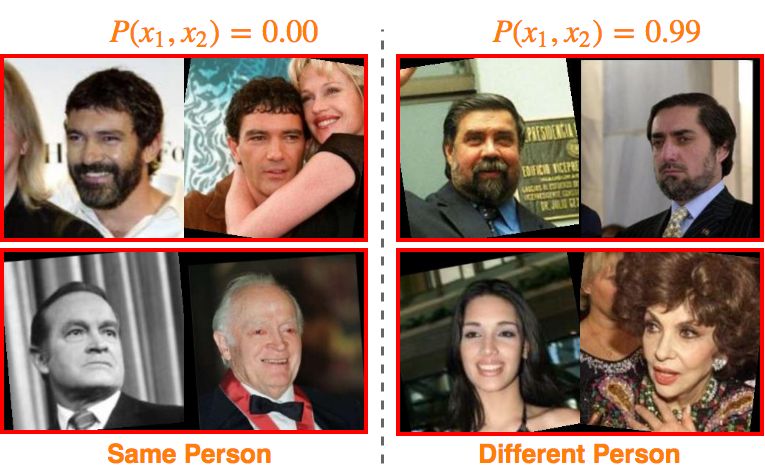}}\hfill
\subfloat[Successful Cases]{\includegraphics[width = 0.49\columnwidth]{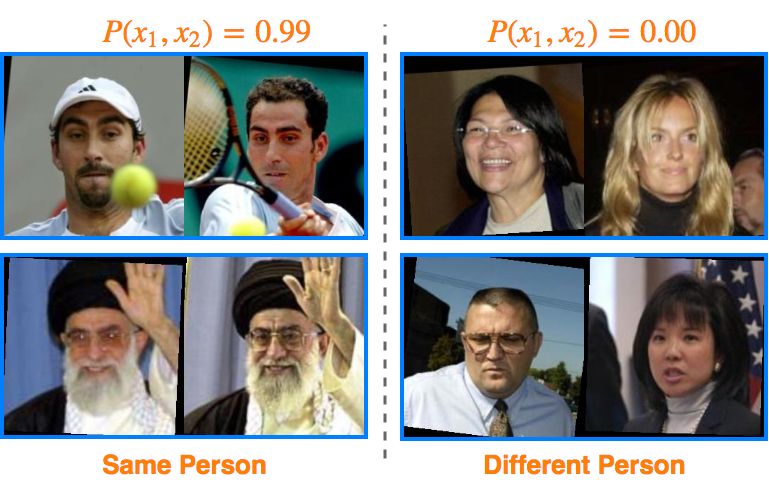}}
\caption{Face verification examples in the test set (View 2) ($\mathcal{T}=0.55$). For images belonging to same person, $P(\bm{x_1}, \bm{x_2})$ should be close to $1$ and for images belonging to different people, $P(\bm{x_1}, \bm{x_2})$ should drop to $0$.}
\vspace{-4mm}
\label{fig:fv_cases}
\end{figure}

The ROC curves of all competing methods are given in Figure~\ref{fig:roc} with corresponding AUC values calculated. It can be observed that the proposed approach \sysname{} outperforms all baseline approaches by providing significantly higher AUC value. As demonstrated in Figure~\ref{fig:roc_weight}, \sysname{} is better mainly because 
it automatically adapts the model complexity to incorporate non-linear models, which help to separate instances that are difficult to distinguish in simple linear models. Figure~\ref{fig:fv_cases} presents some example comparisons in the test set for both successful and failure cases. Here, the comparison probability $P(\bm{x_1}, \bm{x_2})$ computed based on Eq.~\ref{eq:sim_com} is expected to approach $1$ if both images are from the same person and should be close to $0$ if they belong to different people. Apparently, although \sysname{} performs much better than baselines, it may still fail in some cases where severe concept drift, e.g. aging and shave, are present.

\subsection{Image Retrieval}
\subsubsection{Dataset and Experiment Setup}
To demonstrate the performance of \sysname{} on image retrieval task, we show experiments on two benchmark datasets, which are CARS-196\footnote{https://ai.stanford.edu/~jkrause/cars/car\_dataset.html}~\cite{KrauseStarkDengFei-Fei} and CIFAR-100\footnote{https://www.cs.toronto.edu/~kriz/cifar.html}~\cite{Krizhevsky09learningmultiple}. 
The CARS-196 dataset is divided into $8,144$ training images and $8,401$ test images, where each class has been split roughly in a 50-50 ratio. 
Similarly, we randomly split CIFAR-100 into $30,000$ training images and $30,000$ test images by equally dividing images of each class. 
For each image in CARS-196 and CIFAR-100, its deep feature, i.e., the output of the last convolutional layer, is extracted from a VGG-19 model pretrained on  the ImageNet~\cite{ILSVRC15} dataset. 
Details of these 2 datasets are listed in the ``Image Retrieval'' part of Table~\ref{tab:dataset}. 
We form the development set of each dataset by including all its training images.
For each dataset, hyper-parameters of baseline approaches were set based on values reported by the authors, and fine-tuned via 10-fold cross-validation on the development set. In \sysname{}, we set $\beta=0.99$, $L=5$, $S_{hidden}=100$, $s=0.1$, $\eta=0.3$ and $S_{emb}=50$ as default. The value of $\tau$ in \sysname{} was found via 10-fold cross-validation on the development set. 

\subsubsection{Constraint Generation and Evaluation Metric}
With the training images in the development set, we generate the training constraints in the same way as Section~\ref{subsubsec: escg}. However, in this experiment, we sample $50,000$ constraints ($25,000$ seeds) for each dataset due to the vast amount of classes included in them. The Recall@K metric~\cite{JegouDS11} is applied for performance evaluation. Each test image (query) first retrieves K most similar images from the test set and receives a 
score of 1 if an image of the same class is retrieved among these K images and 0 otherwise. Recall@K averages this score over all the images.

\subsubsection{Analysis}

\begin{figure}[t]
\centering
\subfloat[\label{fig:recall_cars}]{\includegraphics[width =0.5 \columnwidth]{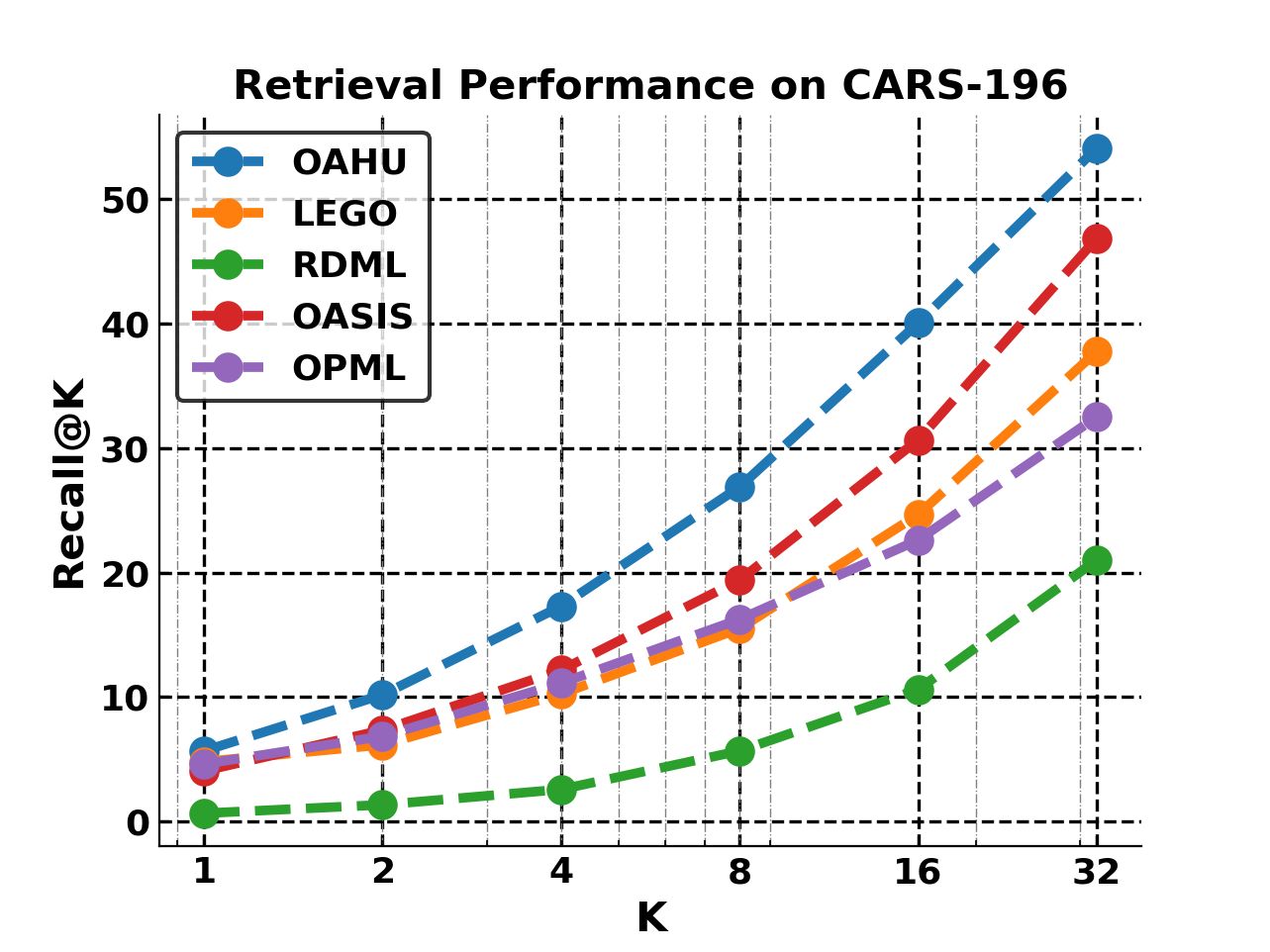}}\hfill
\subfloat[\label{fig:weight_cars}]{\includegraphics[width = 0.5\columnwidth]{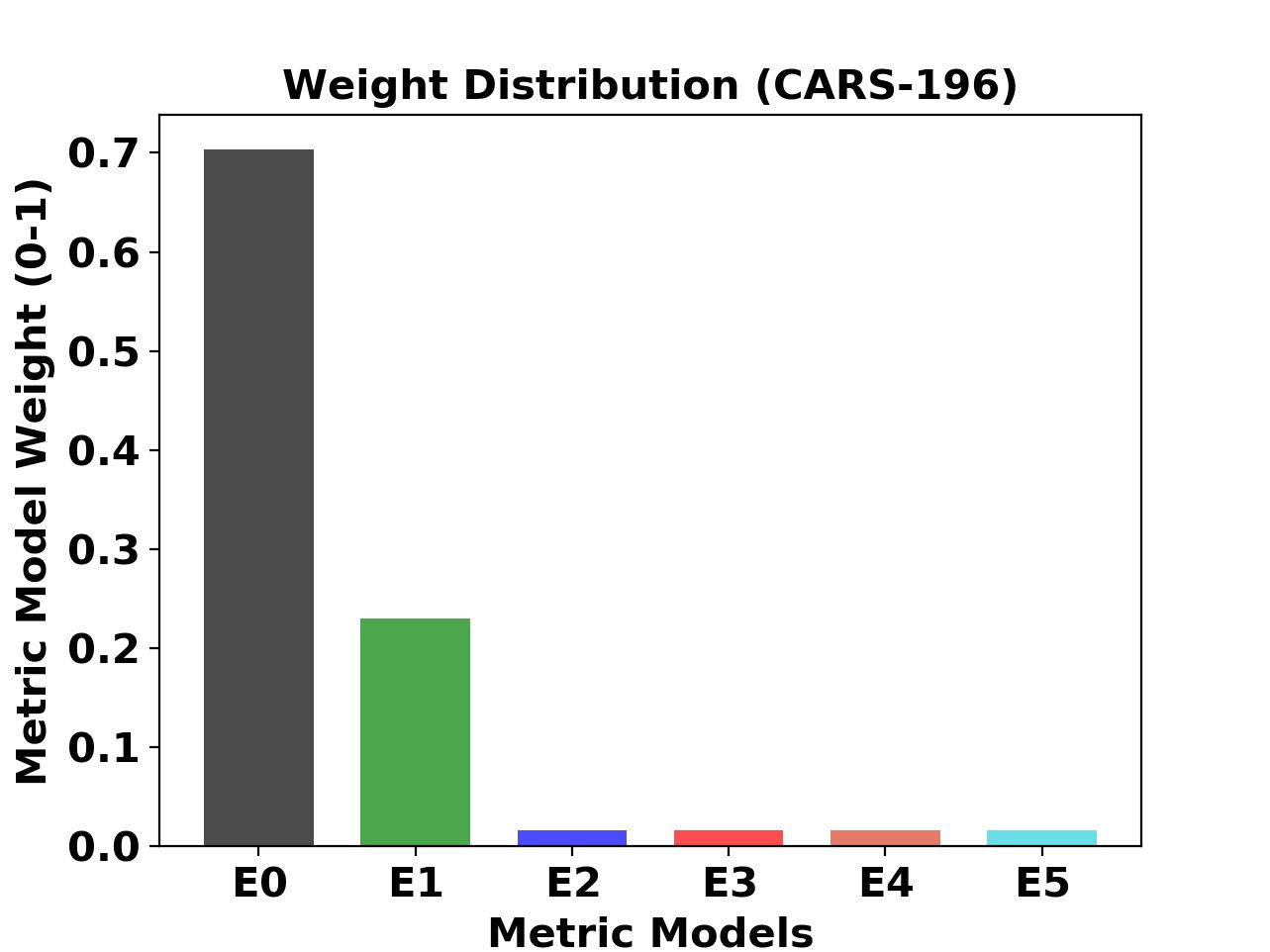}}
\caption{(a) Recall@K score on the test split of CARS-196. (b) Metric weight distribution of \sysname{} on CARS-196.}
\vspace{-4mm}
\label{fig:img_retr_cars}
\end{figure}

\begin{figure}[t]
\centering
\subfloat[\label{fig:recall_cifar100}]{\includegraphics[width =0.5 \columnwidth]{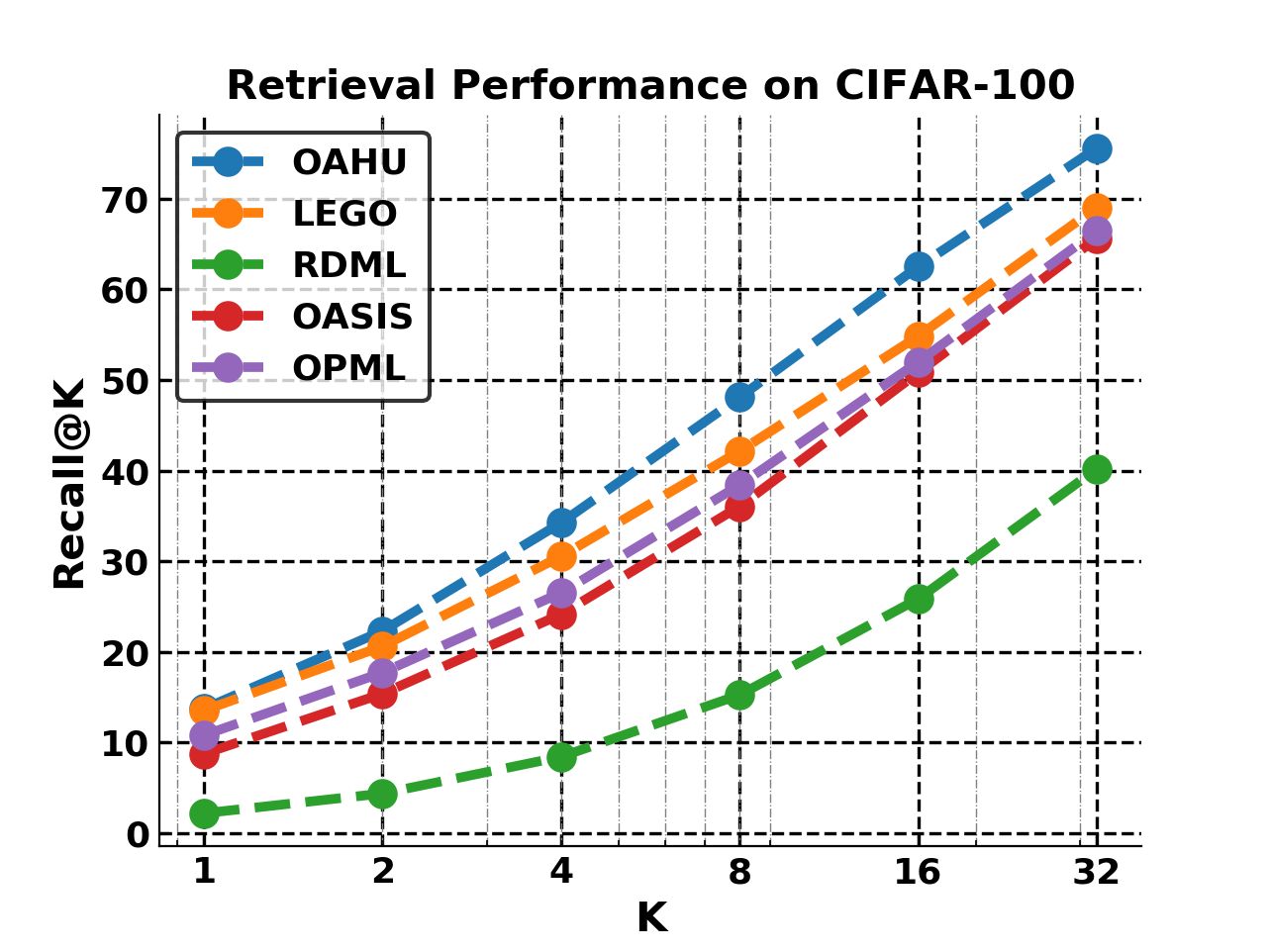}}\hfill
\subfloat[\label{fig:weight_cifar100}]{\includegraphics[width = 0.5\columnwidth]{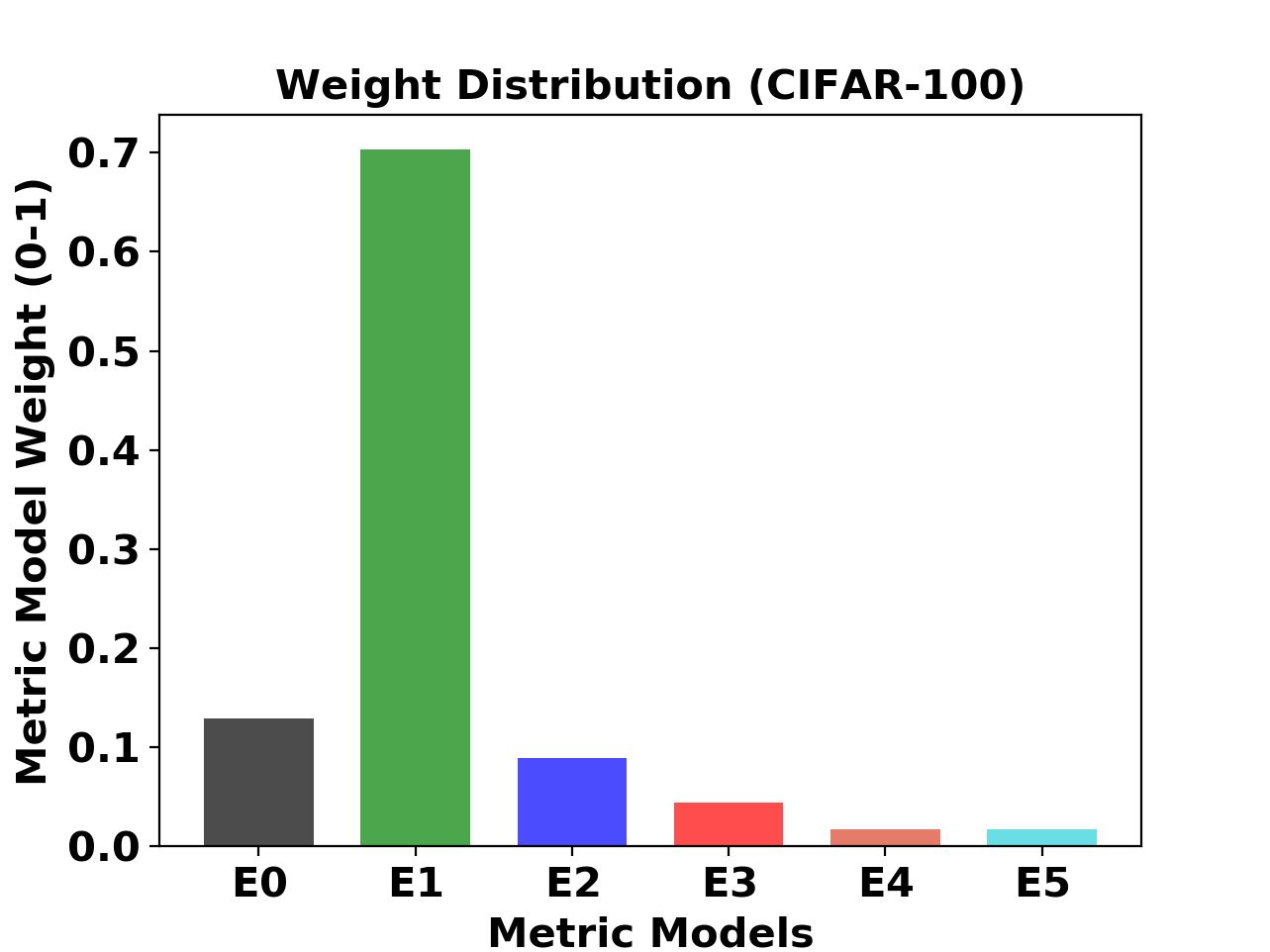}}
\caption{(a) Recall@K score on the test split of CIFAR-100. (b) Metric weight distribution of \sysname{} on CIFAR-100.}
\vspace{-4mm}
\label{fig:img_retr_cifar100}
\end{figure}

\begin{figure}[t]
\centering
\subfloat[CARS-196]{\includegraphics[width = 0.49\columnwidth]{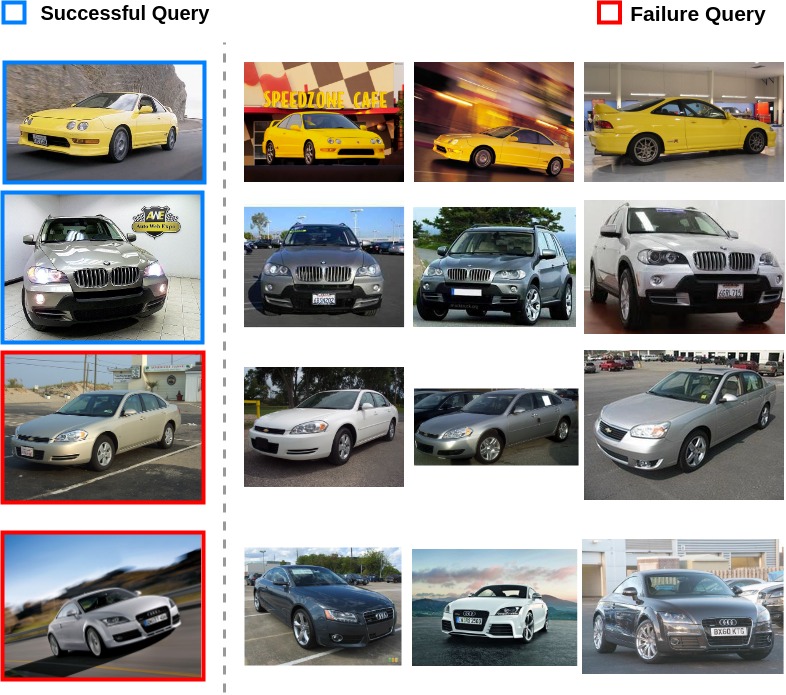}}\hfill
\subfloat[CIFAR-100]{\includegraphics[width = 0.49\columnwidth]{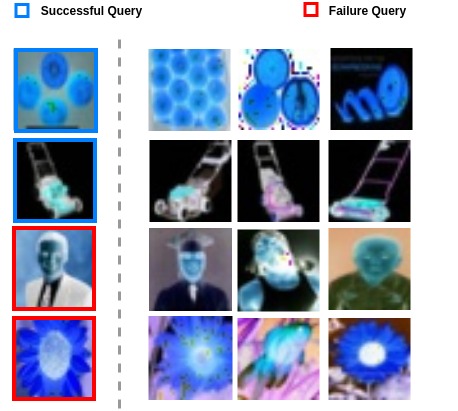}}
\caption{Examples of successful and failure queries on the CARS-196 and CIFAR-100 datasets using our embedding. Images in the first column are query images and the rest are three most similar images. Best viewed on a monitor zoomed in.}
\vspace{-4mm}
\label{fig:retrieval_cases}
\end{figure}

The Recall@K scores on the test set of CARS-196 and CIFAR-100 are shown in Figure~\ref{fig:recall_cars} and Figure~\ref{fig:recall_cifar100}, respectively. We observe that \sysname{} outperforms all baseline approaches by providing significantly higher Recall@K scores. In addition, the Recall@K score of \sysname{} grows more rapidly than that provided by most baselines. Figure~\ref{fig:weight_cars} and Figure~\ref{fig:weight_cifar100} illustrate the metric weight distribution of \sysname{} on CARS-196 and CIFAR-100 respectively. Compared to the baselines that are only linear models, \sysname{} takes full advantage of non-linear metrics to improve the model expressiveness. Figure~\ref{fig:retrieval_cases} shows some examples of successful and failure queries on both datasets.
Despite the huge change in the viewpoint, configuration, and illumination, our method can successfully retrieve examples from the same class and most retrieval failures come from subtle visual difference among images of different classes.

\subsection{Sensitivity of Parameters}


The three main parameters in \sysname{} is the control parameter $\tau$. 
We vary $\tau$ from 0.1 to 0.9 and $S_{emb}$ from 10 to 100 to study their effect on the classification performance. 
We observe that both classification accuracy (1-error rate) and $F_1$ score significantly drop if $\tau$ is greater than 0.6 but remain unchanged with various $\tau$ values if $\tau\leq 0.6$. This observation verifies Theorem~\ref{theo:1} which states that classes are separated if $\tau \in (0,\frac{2}{3})$. On the other hand, the classification accuracy of \sysname{} peaks at embedding size $S_{emb}=50$ where all semantic information are maintained. If we further increase the embedding size, noisy information is going to be introduced which degrades the classification performance. Therefore, we suggest to choose $\tau=0.1$ and $S_{emb}=50$ as default.

\section{Limitations and Conclusions}
In this paper, we propose a novel online metric learning framework, called \sysname{}, that learns a ANN-based metric with adaptive complexity and a full constraint utilization rate. Compared to the state-of-the-art baselines, \sysname{} not only obtains the best performance by automatically adapting its complexity to improve the model expressiveness, but also is more robust in terms of the quality of input constraints. 
Like existing state-of-the-art solutions, \sysname{} fails when facing severe concept drift data that invalidate the learned similarity metric. 
We leave the extension of \sysname{} to severe concept drift scenarios for future work.
\label{Sec:conclusion}

\section{ACKNOWLEDGMENTS}
This material is based upon work supported by NSF award numbers: DMS-1737978,  DGE 17236021. OAC-1828467; ARO award number: W911-NF-18-1-0249; IBM faculty award (Research); and NSA awards.

\bibliographystyle{ACM-Reference-Format}
\balance
\bibliography{reference}

\end{document}